\documentclass{sn-jnl}% APA Reference Style 
\usepackage[round, authoryear]{natbib}
\usepackage{graphicx}%
\usepackage{multirow}%
\usepackage{amsmath,amssymb,amsfonts}%
\usepackage{amsthm}%
\usepackage{mathrsfs}%
\usepackage[title]{appendix}%
\usepackage{xcolor}%
\usepackage{textcomp}%
\usepackage{manyfoot}%
\usepackage{booktabs}%
\usepackage{listings}%
\usepackage{comment}
\usepackage{subfig}
\usepackage{enumitem}
\usepackage{multirow}
\usepackage{multicol}
\usepackage{array}
\usepackage[ruled]{algorithm2e}

\definecolor{myblue}{RGB}{0, 50, 150}

\newcommand{\bftab}{\fontseries{b}\selectfont} % non extended bold face

\usepackage{background}

\begin{document}

\SetBgContents{Published at \url{https://doi.org/10.1007/s10479-026-07291-x}}      % a watermark
\SetBgPosition{current page.center}
\SetBgAngle{0}                                    % rotate
\SetBgColor{gray}                                 % color
\SetBgScale{1.5}                                  % scale
\SetBgHshift{0}                                   % location x=0 for center
\SetBgVshift{9cm} 

\title[Optimizing accuracy and diversity: a multi-task approach to forecast combinations]{Optimizing accuracy and diversity: a multi-task approach to forecast combinations}


\author[1]{\fnm{Giovanni} \sur{Felici}}\email{giovanni.felici@iasi.cnr.it}

\author*[1,2]{\fnm{Antonio M.} \sur{Sudoso}}\email{antoniomaria.sudoso@uniroma1.it}

\affil[1]{\orgdiv{Institute for Systems Analysis and Computer Science ``Antonio Ruberti''}, \orgname{National Research Council}, \orgaddress{\street{Via dei Taurini 19}, \city{Rome}, \postcode{00185}, \country{Italy}}}

\affil[2]{\orgdiv{Department of Computer, Control and Management Engineering ``Antonio Ruberti''}, \orgname{Sapienza University of Rome}, \orgaddress{\street{Via Ariosto 25}, \city{Rome}, \postcode{00185}, \country{Italy}}}


\abstract{
We present a multi-task optimization approach based on a deep learning architecture for time series forecasting. 
{We leverage large collections of time series to identify the weights of forecasting models that can be combined to produce forecasts for each series.} 
This method jointly addresses two tasks: the selection of different forecasting models, and their effective combination. In doing so, it keeps into account, in an original way, both the accuracy and diversity of the forecasting methods.
For a given time series, the model combination module extracts features and uses them to optimize the weights of the forecasting methods. Simultaneously, the model selection module extracts other features to identify the subset of methods to be used for the prediction. This selection process is framed as a classification problem, with the labels representing the set of models to be used for a series. These labels are determined by solving an auxiliary optimization problem that identifies accurate and diverse methods for each time series.
The outputs of the two modules are then combined and the entire neural network is jointly trained by minimizing a custom loss function via gradient descent optimization.
Experimental results on a large set of series from the M4 competition dataset {and from real road traffic data} show that our proposal enhances point forecast accuracy compared to state-of-the-art methods.
}




\maketitle

%\section*{Statements and Declarations}
%\subsection*{Competing Interests}
%The authors have no financial or non-financial interests to disclose.
%\subsection*{Data Availability}
%The data that support the findings of this study are available from the corresponding author upon request.
%\subsection*{Funding}
%The work presented in this paper has been supported by PNRR MUR project PE0000013-{FAIR} and CNR DIT.AD106.097 project UISH - Urban Intelligence Science Hub.
%\subsection*{Ethical approval}
%This article does not contain any studies with human participants or animals performed by any of the authors.

%\newpage

\section{Introduction}\label{section:intro}
Forecasting is an important and complex task, with applications in the most diverse domains, among which finance \citep{tung2009financial}, energy \citep{repetto2024artificial, sharma2024incorporating}, supply chains \citep{syntetos2016supply, paul2022ensemble}, inventory management \citep{syntetos2009forecasting, efat2024deep}, {traffic and mobility \citep{chrobok2004different, jiang2022graph}}.
Given its interest and complexity, the field of forecasting has frequently leveraged mathematical optimization methodologies, becoming the object of a successful vein of literature within operational research \citep{fildes2008forecasting}. Recently, this trend has been further reinforced by the cross-fertilization between optimization and machine learning literature \citep{gambella2021optimization}.

One of the most successful approaches to forecasting is \textit{forecast combination}, where the results of different forecasting methods are aggregated to produce a single, more accurate prediction \citep{bates1969combination, clemen1989combining, timmermann2006forecast, scholz2022forecast}. Indeed, many studies have shown that forecast combinations lead to improvements in forecast accuracy {as they reduce the variance of the individual forecasts} \citep{atiya2020does}.
%The M4 competition \citep{makridakis2020m4} represents a recent installment within a series of M-competitions \citep{makridakis1982accuracy} that have spanned several decades. These competitions have proven invaluable in enabling researchers to acquire comprehensive insights into the performance of diverse forecasting models and establish best practices in the field \cite{hyndman2020brief}. 
The dominance of forecast combinations over other methods has become evident in the M4 competition \citep{makridakis2020m4}, a well-established benchmark.
%that has proven to be highly valuable for researchers. 
%The competition provides insights into the performance of various forecasting models and facilitates the establishment of best practices in the field \citep{hyndman2020brief}. 
This competition built upon the knowledge gained from previous competitions, confirming that forecast combinations obtains consistently better results on the considered benchmarks. Out of the top 17 most accurate methods, 12 were combinations of forecasting models. 
For a recent overview of forecast combinations, we direct the readers to the review article by {\cite{wang2023forecast}}.
In essence, forecast combinations may take place with different methods with varying levels of complexity.
One of the simplest and most commonly used methods for combining forecasts is taking the arithmetic mean of the individual forecasts \citep{jose2008simple}. 
%This approach is easy to implement and is often effective. 
{This approach is often noted in the literature for outperforming more complex combining methods, a phenomenon known as the ``forecast combination puzzle''}. Consequently, it serves as a benchmark in practical applications against other methods.
More sophisticated techniques include weighted averages, where weights can be determined through regression techniques \citep{winkler1992sensitivity}, or with machine learning and optimization approaches \citep{wang2023forecast}. 

\cite{lichtendahl2020some} conducted a study to investigate why some combinations of methods performed better than others in the %recent 
M4 competition. Their findings highlighted the significance of \textit{diversity} and \textit{accuracy} of the individual models as crucial factors for effective forecasting \citep{atiya2020does}. 
{Indeed, the integrated role of accuracy and diversity has been explored in several statistics and machine learning studies. For instance, the ambiguity decomposition theory \citep{krogh1994neural} and
the bias-variance-covariance decomposition \citep{ueda1996generalization} provide a quantification of accuracy and diversity for convex-combined models through the mean squared error criterion. The same idea has been applied by \cite{kang2022forecast} to the forecast combination task.}

\newpage

Our paper proposes a model-based approach designed to optimize convex combinations of forecasting models, known as \textit{base forecasters}. Let $M$ and $H$ be the number of base forecasters and the forecast horizon, respectively. For a given time series $\{y_1, \dots, y_T\}$ of length $T$, we denote the $h$-th step forecast produced by the $i$-th individual method as $\hat{y}_{i, T+ h}$, where $i = 1, \dots, M$ and $h = 1, \dots, H$. We say that the combined forecast ${\hat{y}_{\textrm{comb}, T+h}}$ at the $h$-th step is a convex combination of $M$ base models where
\begin{align}\label{eq:convex_comb}
    \hat{y}_{\textrm{comb}, T+h} = \sum_{i=1}^{M} w_i \hat{y}_{i, T+h} \quad \textrm{s.t.} \quad \sum_{i=1}^M w_i = 1, \ w_i \geq 0 \quad \forall i \in \{1, \dots, M\},
\end{align}
and $w_i$ is the weight of the forecast produced by the $i$-th method. {Convex combinations are easy to interpret since the weights are constrained to be non-negative and sum up to one}, making it clear how much influence each forecast has on the final outcome \citep{wang2023forecast}.
Given this general framework, building a forecast combination model involves two steps: selecting the base forecasters and choosing an appropriate method for combining their outputs.

Recent literature uses the term \textit{meta-learning} to describe the process of automatically acquiring knowledge for model selection or combination \citep{prudencio2004meta, lemke2010meta}.
{When forecasting large collections of time series, meta-learning may be adopted to obtain the most appropriate model or the optimal combination of candidate models, per series. Such learning is based on special features extracted from the time series that provide a meaningful representation of the characteristics and patterns in the series}.
%
%The concept of meta-learning for time series forecasting is by \cite{talagala2018meta}, where a machine learning model is employed to establish connections between time series features and the forecasting performance of base forecasters and finally identify the ideal forecaster for a given time series.
The method we propose is based on meta-learning and uses deep neural networks and optimization to select and combine the available forecasting methods by simultaneously exploiting accuracy and diversity criteria. %Then, the base forecasters are combined with a convex linear function.
{The idea of taking into account diversity and accuracy is not new in the forecasting literature but is indeed a novelty in the context of meta-learning for time series forecasting. For instance, \cite{diebold2019machine} propose a methodology that jointly performs forecast combination and selection. Specifically, they employ a LASSO-type regression approach which sets some combining weights to zero and shrinks the remaining weights toward equality. This method determines the weights for each time series independently; differently, our approach leverages large collections of time series to identify the appropriate weights for any given series by considering many other series.}

%To advance the notation used in the remainder of the paper, %This transparency allows decision-makers to understand the relative importance of each forecast in shaping the overall prediction \citep{wang2023forecast}.}

The remainder of the paper is organized as follows. Section \ref{section:related} discusses related meta-learning methods proposed in the literature and outlines the main contributions of the paper. %Section \ref{section:forecasters} presents the evaluation metrics and a description of the employed pool of forecasting methods. 
Section \ref{section:methodology} outlines the proposed multi-task optimization methodology. Section \ref{section:implementation} {describes the datasets, the evaluation metrics, and the implementation details}.
Section \ref{section:experimental} describes the experimental validation. Finally, Section \ref{section:conclusion} concludes the paper with possible future research directions.

\section{{Related works:} forecasting by meta-learning}
\label{section:related}
%Meta-learning approaches use a group of time series to establish connections between the characteristics of the time series and the out-of-sample performance of various forecasting models
Meta-learning can be employed to either select the most appropriate model or to optimize the weights used to combine the different forecasting methods. 
\cite{talagala2018meta} developed a meta-learning approach called FFORMS (Feature-based FORecast Model Selection), which uses a Random Forest (RF) to select the best forecasting method from nine base forecasters, based on a set of manually selected time series features. 
%To build a reliable classiﬁer, they proposed augmenting the set of observed time series by simulating new time series similar to those in the assumed population. 
%Time series features are manually selected in the number of 25 for non-seasonal data and 30 for seasonal data. 
\cite{montero2020fforma} improved FFORMS and proposed a meta-learning approach to learn the weights of convex combinations of forecasting models, resulting in the framework called FFORMA (Feature-based FORecast Model Averaging). 
In FFORMA, before performing the actual forecast, 42 hand-crafted features are extracted from the original time series. To determine the optimal weights of the combination, the problem is framed as a non-linear regression where time series features are linked with the forecasting errors using the XGBoost algorithm \citep{chen2016xgboost}. %FFORMA placed second in the M4 competition \citep{makridakis2020m4}.
\cite{di2022sparse} proposed a meta-learning system based on a Multilayer Perceptron (MLP) that takes as input the same pre-computed time series feature used in FFORMA and automatically provides sparse convex combinations of the methods. One advantage of this approach is that it avoids to compute forecasts for the excluded methods, leading to computational savings. However, the obtained results are worse in terms of accuracy than the approach by \cite{montero2020fforma}. 

The relative importance of accuracy and diversity in the selection of combination methods is of great interest to the community; solid conclusions on the issue are found in \cite{lichtendahl2020some}, where a detailed analysis of the top combination method of the M4 competition is performed. They found that a simple trimmed mean of a subset of methods was nearly as effective as the combination produced by \cite{montero2020fforma}.
To incorporate diversity, \cite{kang2022forecast} tailored the FFORMA framework to allow for the diversity of the forecasts as the inputs. This results in a supervised approach where time series features are extracted by looking at the diversity of forecasts among the methods in the pool. %The newly obtained hand-crafted features are then employed to train the weighted combination of forecasting models. 
The inclusion of diverse methods within the combination scheme enhanced the forecasting accuracy of FFORMA. 

In the methods mentioned above, time series features are identified a priori from a set of established ones. Another body of research focuses on the use of Deep Neural Networks (DNN) to extract features in a data-driven fashion. For instance, \cite{ma2021retail} proposed a meta-learning algorithm designed for retail forecasting where the features are extracted using a Convolutional Neural Network (CNN), and then linked with weights that combine the forecasting methods. \cite{li2020forecasting} adopted a similar approach, where a CNN is employed to learn time series features from images of the recurrence plots, leveraging computer vision algorithms. Table \ref{tab:meta-literature} offers a comparison of the recent research studies on time series forecasting with meta-learning. 
\begin{table}[!ht]
    \centering
    \footnotesize
    \begin{tabular}{lllll}
    \toprule
\textbf{Paper}&\textbf{Feature extr.}&\textbf{Meta-learner}&\textbf{Optimization task}&\textbf{Diversity}\\
\midrule
\multirow{2}{*}{\cite{talagala2018meta}} & Judgmental & \multirow{2}{*}{RF} & \multirow{2}{*}{{Selection}} & \multirow{2}{*}{No}\\
& Unsupervised & & & \\
\midrule
\multirow{2}{*}{\cite{montero2020fforma}} & Judgmental & \multirow{2}{*}{XGBoost} & \multirow{2}{*}{{Combination}} & \multirow{2}{*}{No}\\
& Unsupervised & & & \\
\midrule
\multirow{2}{*}{\cite{li2020forecasting}} & Automatic & \multirow{2}{*}{DNN} & \multirow{2}{*}{{Combination}} & \multirow{2}{*}{No}\\
& Supervised & & & \\
\midrule
\multirow{2}{*}{\cite{lichtendahl2020some}} & Judgmental & \multirow{2}{*}{XGBoost} & \multirow{2}{*}{{Combination}} & Yes\\
& Unsupervised & & & Post-processing\\
\midrule
\multirow{2}{*}{\cite{ma2021retail}} & Automatic & \multirow{2}{*}{DNN} & {Selection or} & \multirow{2}{*}{No}\\
& Supervised & & {combination} &\\
\midrule
\multirow{2}{*}{\cite{di2022sparse}} & Judgmental & \multirow{2}{*}{MLP} & {Selection or} & \multirow{2}{*}{No}\\
& Unsupervised & & {combination} &\\
\midrule
\multirow{2}{*}{\cite{kang2022forecast}} & Judgmental & \multirow{2}{*}{XGBoost} & \multirow{2}{*}{{Combination}} & Yes\\
& Supervised & & & Pre-processing\\
\midrule
\multirow{2}{*}{This paper} & Automatic & \multirow{2}{*}{DNN} & {Combination with} & Yes\\
& Supervised & & {auxiliary supervision} & Learned\\
\bottomrule
    \end{tabular}
    \caption{Research on time series forecasting with meta-learning. In the ``Feature extr.'' column, ``judgmental'' denotes manually crafted time series features, while ``automatic'' signifies features identified by the meta-learner. The ``Diversity'' column assesses whether the meta-learner accounts for diversity, and how.}
    \label{tab:meta-literature}
\end{table}
\newline\newline
\noindent
\textbf{Contributions.} Building upon existing literature, we use meta-learning to optimize the weights of convex combinations of forecasting methods. We cast this optimization problem as a {multi-task learning} (MTL) problem. MTL 
solves related tasks simultaneously with the goal of improving the performance of each of them \citep{caruana1997multitask}. 
{In our approach, the main task is forecast combination, while an auxiliary task provides diversity-aware supervision during training.}
%In our approach we address two tasks: forecast combination and forecast selection. 
To this end, we design a deep neural network with two branches, one for each task. The outputs of these branches are combined and the entire network is trained by minimizing a custom loss function. We empirically demonstrate the effectiveness of the approach by testing it on a large number of series from the M4 competition and {on a large-scale traffic flow time series dataset.}
The proposed approach contributes to this stream of literature mainly in three aspects (see also Table \ref{tab:meta-literature}):

\begin{enumerate}
    \item \textit{Optimization task}: 
    The meta-learner addresses the forecast combination problem through a multi-task architecture in which the main task, formulated as a regression problem, learns the weights that minimize the combined forecasting error. {In addition, an auxiliary classification task provides supervision by identifying subsets of forecasting methods that are jointly accurate and diverse. These subsets are obtained by solving an auxiliary optimization problem, whose output is used to generate training labels for the auxiliary task. Rather than enforcing hard model selection, the auxiliary task acts as a diversity-aware regularizer that guides the regression branch toward combinations involving diverse and accurate forecasters};

    \item \textit{Diversity}: In the literature, diversity among the base forecasters has been considered either as a pre-processing or post-processing step  (\cite{lichtendahl2020some}, \cite{kang2022forecast}). 
    {Our approach instead incorporates diversity directly during training by leveraging the auxiliary optimization problem to generate supervision signals}. In this way, diversity is not imposed externally but learned as part of the model training process, influencing the resulting forecast combinations;
    
    \item \textit{Feature extraction}: We compensate for the limited interpretability of deep networks with gradient-based visual explanations, a technique that indicates the most discriminative regions in the input series.

\end{enumerate}

\section{Meta-learner: forecast combination and selection tasks}\label{section:methodology}
As anticipated, our meta-learning approach aims to determine a set of weights that combine forecasts generated from a pool of methods, with the goal of exploiting accuracy and diversity among these methods. 
{To better elaborate on the motivation of our approach}, we consider as a counterpart the work of \cite{talagala2018meta}, where meta-learning is used for selecting the single best method for each series based on the smallest forecasting error. This transforms the problem into a traditional classification problem, where the individual forecasting methods are encoded as classes and the best method becomes the target class for each time series. 
However, there might be other methods that yield similar forecast errors to the best method, making the specific class chosen less relevant compared to the forecast error produced by each method; this leads to the forecast combination approach by \cite{montero2020fforma}, framed as a regression problem, where the weights of the different methods are found by minimizing the error of the combined forecasts. The regression approach can be seen as a classification problem with varying per-class weights for each instance, implying that the nature of the combination task is closely related to the selection one. In the following, we describe how to exploit the interplay between these two tasks within a multi-task optimization methodology.

The proposed meta-learning framework consists of two distinct phases: meta-data generation \& model training (offline phase), and forecasting (online phase). 
In the offline phase, each time series is divided into training and testing periods, where the length of the testing period matches its forecasting horizon. For each time series, {the meta-data generating process involves fitting each forecasting method on the training period and extracting the forecasts produced by the different methods on the testing period.} The forecasts from different methods are gathered into a matrix and then compared with the actual observations of the test period, leading to the matrix of forecasting errors. From this matrix, accuracy and diversity information are summarized as {classification} labels by solving an auxiliary optimization problem.
Subsequently, a meta-learner implemented by a deep neural network is trained using gradient descent optimization, minimizing a custom loss function. This step aims to estimate combination weights for each series, allowing the production of weights and hence the combined forecasts for any target series in the online phase.

\subsection{Neural network design}
In our multi-task methodology, the meta-learning model is a deep neural network composed of two subnetworks, or branches: the first one solves a regression problem while the second one solves a classification problem. The goal of the regression problem is to learn the weights of the base forecasting methods by minimizing the error of combined forecasts, while {an auxiliary task provides diversity-aware supervision through a classification problem}.
The outputs of the two tasks are then combined to obtain the final weights of the convex combination, {with the auxiliary branch acting as a gating mechanism on the regression outputs. We note that the regression branch alone can in principle assign low weights to unsuitable forecasters, thereby performing an implicit form of soft selection. However, this mechanism is driven solely by the forecast-combination loss and does not explicitly account for relationships among forecasting methods, such as redundancy due to high similarity. The auxiliary branch therefore provides an additional signal that is not captured by the forecast-combination loss alone. This signal is incorporated through the classification loss and biases the learned combination weights toward subsets of forecasters that are jointly accurate and diverse}.

We consider a collection of $N$ time series $\{\mathbf{s}^1, \dots, \mathbf{s}^N \}$ of length $T$, a set of $M$ forecasting methods and a forecasting horizon of length $H$. Every time series $\mathbf{s}^i$ is split into training $\mathbf{x}^i = [y^i_{1}, \dots, y^i_{T}]^\top$ and test $\mathbf{y}^i = [y^i_{T+1}, \dots, y^i_{T+H}]^\top$ periods. We denote by ${\mathbf{F}}^i \in \mathbb{R}^{H \times M}$ the matrix of forecasts produced by the $M$ methods for the entire forecasting horizon $H$, where ${F}^i_{hm}$ is the $h$-step ahead forecast produced by the method $m \in \{1, \dots, M\}$ for the series $i \in \{1, \dots, N\}$. Let $\mathbf{1}_M$ be the {$M$-dimensional vector of all ones}, then $\mathbf{E}^i = \mathbf{y}^i \mathbf{1}_M^\top - {\mathbf{F}}^i$, where $E^i_{hm} = y^i_{T+h} - F^i_{hm}$ for all $h \in \{1, \dots, H\}$ and $m \in \{1, \dots, M\}$, represents the matrix of forecasting errors produced by the $M$ methods over the forecasting horizon $H$ for the series $i \in \{1, \dots, N\}$.

The regression subnetwork takes as input raw time series $\mathbf{x}^i$ of length $T$, extracts time series features by means of convolutional layers, and returns a set of $M$ un-normalized weights.
We denote the un-normalized weights estimation model $f_{\textrm{reg}} \colon \mathbb{R}^T \to \mathbb{R}^M$ as 
\begin{equation*}
    \hat{\mathbf{o}}^i_{\textrm{reg}} = f_{\textrm{reg}}(\mathbf{x}^i; \mathbf{\theta}_{\textrm{reg}}).
\end{equation*}
Thus, the function $f_{\textrm{reg}}$ is a function parameterized by the subnetwork weights $\mathbf{\theta}_{\textrm{reg}}$ which first maps a time series $\mathbf{x}^i  \in \mathbb{R}^T$ to a set of extracted features $\mathbf{h}_{\textrm{reg}}^i \in \mathbb{R}^D$, where $D$ is the dimension of the learned feature vector, and then outputs a set of un-normalized weights $\hat{\mathbf{o}}^i_{\textrm{reg}}  \in \mathbb{R}^M$ of the base forecasters for that time series. 

To learn the most appropriate base forecasters for a specific time series, we adopt a multi-label classification approach, where the individual forecasting methods are encoded as classes and the most accurate and diverse methods become the target classes for each time series. To generate training labels for the classification task we adapt the Quadratic Programming (QP) feature selection method proposed by \cite{rodriguez2010quadratic}.
Given a set of $M$ forecasting methods, our goal is to assign higher importance to methods satisfying two key requirements: accuracy and diversity among the methods.
The resulting optimization problem for the $i$-th time series can be expressed as:
\begin{equation}
\begin{aligned}\label{prob:selection}
\min_{\mathbf{v}} \quad & \frac{1}{2}(1-\alpha^i) \mathbf{v}^\top \mathbf{Q}^i \mathbf{v} - \alpha^i \mathbf{v}^\top \mathbf{c}^i\\
\textrm{s.t.} \quad & \mathbf{1}_{M}^\top \mathbf{v} = 1, \ \mathbf{v} \geq \mathbf{0}_M.
\end{aligned}\tag{QP-LAB}
\end{equation}
where {$\mathbf{v}$} is an $M$ dimensional vector representing the relative importance of each method, $\mathbf{Q}^i$ is an $M \times M$ symmetric positive semidefinite matrix that captures the pairwise similarities among the forecasting methods, 
{with higher values indicating greater similarity and lower values indicating greater diversity}, and $\mathbf{c}^i$ is an $M$ dimensional vector representing the accuracy of each forecasting methods for the time series $i \in \{1, \dots, N\}$. 
{Thus, Problem \eqref{prob:selection} aims at minimizing the similarity among the methods (quadratic term) while maximizing their accuracy (linear term)}. 
The scalar quantity $\alpha^i \in [0, 1]$ controls the trade-off between similarity and accuracy. {To balance the quadratic and linear terms in the objective function, we set $\alpha^i = \frac{\bar{q}^i}{\bar{q}^i + \bar{c}^i}$, which satisfies the condition $(1-\alpha^i)\bar{q}^i = \alpha^i \bar{c}^i$ and ensures that the two terms contribute equally on average. Here, $\bar{q}^i$ is the mean value of the elements of the matrix $\mathbf{Q}^i$ and $\bar{c}^i$ is the mean value of the elements of vector $\mathbf{c}^i$. This results in a data-driven and series-specific choice of $\alpha^i$.}
{If the forecasting methods exhibit high diversity, that is, they have low similarity with each other relative to their accuracy ($\bar{c}^i >\!\!> \bar{q}^i$), then the linear term in \eqref{prob:selection} can dominate the quadratic one. Thus, this choice of $\alpha^i$ reduces this dominance by making $\alpha^i$ small. On the other hand, if the forecasting methods have high similarity relative to their accuracy ($\bar{q}^i >\!\!> \bar{c}^i$), the quadratic term in \eqref{prob:selection} can dominate the linear one. In this case, our choice of $\alpha^i$ makes it larger and thus the objective function becomes more balanced. Details on how $\mathbf{Q}^i$ and $\mathbf{c}^i$ are computed are reported in Section \ref{sec:meta-data-generation}.}
Problem \eqref{prob:selection} is convex since the correlation matrix $\mathbf{Q}^i$ is positive semidefinite. 
By solving this optimization problem for each time series, we identify a set of forecasting methods characterized by a favorable trade-off between accuracy and diversity, which is used to provide supervision to the classification subnetwork. The {ground-truth (i.e., the classification labels)} $\mathbf{o}^i \in \{0, 1\}^M$ for the $i$-th time series is then constructed from the optimal solution $\mathbf{v}^\star$ as follows:
\begin{align*}
o^i_{j} = 
\begin{cases}
    1, & \text{if } v^\star_j \geq \tau \\
    0, & \text{otherwise}
\end{cases} \qquad \forall j \in \{1, \dots, M\},
\end{align*}
where $\tau$ is a user-defined threshold. {While the optimal solution $\mathbf{v}^\star$ provides a continuous characterization of the trade-off between accuracy and diversity, directly learning this target may be challenging due to its sensitivity to small variations in forecasting errors across time series. In contrast, the thresholding step extracts the most relevant information from $\mathbf{v}^\star$, namely the identification of subsets of forecasters that achieve a favorable balance between accuracy and diversity.}
Note that, if $\tau$ is too large, the number of target forecasting methods may decrease up to the point where only one forecasting method is selected. This turns the multi-label classification problem into a conventional single-label one, thereby overlooking the benefits of forecast combinations. In general, the threshold parameter $\tau$ can be tuned via cross-validation {on a validation subset that is separated from the held-out test data used for final evaluation}, with a focus on out-of-sample forecasting accuracy. Moreover, if some base forecasters are not sufficiently represented, one could tune $\tau$ by using various balancing techniques, such as resampling methods or cost-sensitive loss functions \citep{charte2015addressing, tarekegn2021review}. 
{We emphasize that this thresholding step is used to construct supervision for the auxiliary task and does not enforce hard selection at inference time. Rather, it provides a simplified representation of the diversity-aware solution that is easier to learn within a classification framework.}
The auxiliary classification network $f_{\textrm{cls}} \colon \mathbb{R}^T \to [0, 1]^M$ is trained to imitate the decisions of the QP model by learning the mapping
\begin{equation*}
    \hat{\mathbf{o}}^i_{\textrm{cls}} = f_{\textrm{cls}}(\mathbf{x}^i; \mathbf{\theta}_{\textrm{cls}}).
\end{equation*}

The function $f_{\textrm{cls}}$ is parameterized by the subnetwork weights $\mathbf{\theta}_{\textrm{cls}}$.
Parallel to the regression subnetwork, it first maps a time series $\mathbf{x}^i \in \mathbb{R}^T$ to a time series future vector $\mathbf{h}_{\textrm{cls}}^i \in \mathbb{R}^D$ and then outputs a set of predicted labels $\hat{\mathbf{o}}^i_{\textrm{cls}} \in \mathbb{R}^M$. More in detail, each element in $\hat{\mathbf{o}}^i_{\textrm{cls}}$ is a continuous value between 0 and 1 and represents the probability that a method is appropriate for forecasting the $i$-th time series based on accuracy and diversity principles. 

Finally, we apply the softmax function to the combined output of the subnetworks which is obtained by multiplying the output of both branches as
\begin{equation*}
    \hat{{w}}^i_j = \textrm{softmax}_j( \hat{\mathbf{o}}^i_{\textrm{reg}} \odot  \hat{\mathbf{o}}^i_{\textrm{cls}}) = \frac{\textrm{exp}\left((\hat{\mathbf{o}}^i_{\textrm{reg}} \odot  \hat{\mathbf{o}}^i_{\textrm{cls}})_j\right)}{\sum_{k=1}^M \textrm{exp}\left(( \hat{\mathbf{o}}^i_{\textrm{reg}} \odot  \hat{\mathbf{o}}^i_{\textrm{cls}})_k\right)}, \quad \forall j \in \{1, \dots, M\},
\end{equation*}
where $\odot$ represents the element-wise multiplication. The softmax function maps vectors from the Euclidean space to probability distributions, thus allowing to output the estimated weights $\hat{\mathbf{w}}^i \in \mathbb{R}^{M}$ of the convex combination for the $i$-th time series.
\begin{comment}
For a real vector $\mathbf{z} \in \mathbb{R}^{M}$ the softmax function is defined as
\begin{align*}
    \textrm{softmax}_i(\mathbf{z}) = \frac{\textrm{exp}(z_i)}{\sum_{j=1}^M \textrm{exp}(z_j)}, \quad \forall i \in \{1, \dots, M\}.
\end{align*}
\end{comment}
{The element-wise multiplication enables interaction between the regression and auxiliary branches by gating the regression scores according to the outputs of the auxiliary task. Importantly, this mechanism does not enforce hard selection. Instead, each component $(\hat{\mathbf{o}}^i_{\textrm{cls}})_j$ acts as a soft scaling factor that reflects the likelihood that the $j$-th forecasting method belongs to a diversity-aware candidate set. As a result, the final softmax allocation is biased toward forecasters that are jointly accurate and diverse, while the regression branch retains control over the relative magnitudes of the combination weights.}

\subsection{Loss function design}
Generally, for optimizing multi-task learning architectures, it is necessary to weight the importance of each task. Specific to our problem, the combination task represents the main task and hence should receive more attention, whereas the selection task is treated as an auxiliary task to enhance the performance of the main task. Given this requirement, we now present the components of the loss function used for training the overall meta-learner.

As pointed out by \cite{ma2021retail}, one limitation of FFORMA is that it focuses on minimizing the combined errors of the base forecasters, not the combined forecasts directly, which can lead to suboptimal combinations. 
Let $\mathbf{\Theta} = \{\mathbf{\theta}_{\textrm{reg}}, \mathbf{\theta}_{\textrm{cls}}\}$ be the the parameters of the overall neural network. The meta-learner is trained by minimizing the scaled Mean Absolute Error (MAE) with respect to $\mathbf{\Theta}$, which is defined as
\begin{align}
    \mathcal{L}_{\textrm{comb}}(\mathbf{x}^i, \mathbf{y}^i, {\mathbf{F}}^i; \mathbf{\Theta}) &= \frac{1}{N} \sum_{i=1}^{N} \frac{\| \hat{\mathbf{y}}^i - \mathbf{y}^i \|_1}{\| \bar{\mathbf{y}}^i - \mathbf{y}^i \|_1} = \frac{1}{N}\sum_{i=1}^{N} \frac{\| {\mathbf{F}}^i \hat{\mathbf{w}}^i - \mathbf{y}^i \|_1}{\big\| \frac{1}{M} {\mathbf{F}}^i \mathbf{1}_M - \mathbf{y}^i \big\|_1}, 
\end{align}
where $\| \cdot \|_1$ is the $\ell_1$ norm, $\hat{\textbf{y}}^i = {\mathbf{F}}^i \hat{\mathbf{w}}^i = [\hat{y}^i_{T+1}, \dots, \hat{y}^i_{T+H}]^\top \in \mathbb{R}^H$ contains the weighted combination of the forecasts, and $\bar{\mathbf{y}}^i = \frac{1}{M} {\mathbf{F}}^i \mathbf{1}_M$ is the forecast combination obtained by averaging the forecast produced by the $M$ methods. This scaling is useful in the training phase to perceive the differences between the performance of forecast combinations provided by the meta-learner and the simple average combinations which represent a strong baseline. Note that, $\mathcal{L}_{\textrm{comb}}$ is jointly decided by $\hat{\mathbf{o}}^i_{\textrm{reg}}$ and $\hat{\mathbf{o}}^i_{\textrm{cls}}$. {We provide the details of gradient computation in Appendix \ref{appendix:gradient}, illustrating how the outputs of the combination and selection tasks interact and influence the parameter updates of the overall network through back-propagation.}

The classification subnetwork is trained to predict the output labels we are ultimately interested in, and naturally induces the following loss function:
\begin{align}
    \mathcal{L}_{\textrm{cls}}(\mathbf{x}^i, \mathbf{o}^i_{\textrm{cls}}; \mathbf{\Theta}) = -\frac{1}{N} \sum_{i=1}^N \sum_{j=1}^M \Big( (\mathbf{o}^i_{\textrm{cls}})_j \log (\hat{\mathbf{o}}^i_{\textrm{cls}})_j + (1 - (\mathbf{o}^i_{\textrm{cls}})_j)\log (1 - (\hat{\mathbf{o}}^i_{\textrm{cls}})_j) \Big),
\end{align}
that is the binary cross-entropy loss for a multi-label classification problem and should be minimized with respect to the overall neural network weights. 

The final cost function is given by the sum of the cost functions of the combination task and the selection task. The goal of training is to minimize the following loss with respect to the network's parameters $\mathbf{\Theta}$:
\begin{align}
\label{eq:loss}
    \mathcal{L}(\mathbf{x}^i, \mathbf{y}^i, {\mathbf{{F}}}^i, \mathbf{o}^i_{\textrm{cls}}; \mathbf{\Theta}) & = \mathcal{L}_{\textrm{comb}}(\mathbf{x}^i, \mathbf{y}^i, {\mathbf{{F}}}^i; \mathbf{\Theta}) + \lambda \ \mathcal{L}_{\textrm{cls}}(\mathbf{x}^i, \mathbf{o}^i_{\textrm{cls}}; \mathbf{\Theta}),
\end{align}
where $\lambda$ is a hyperparameter chosen by cross-validation to balance the relative contributions of the two tasks in the loss. Increasing $\lambda$ emphasizes the selection task by giving more importance to the classification loss term. {
Note that, the output of the auxiliary branch also enters the combination process through the gating mechanism, influencing the final weights used in the forecast-combination loss. As a result, the diversity-aware information encoded in the auxiliary task affects the learning process both directly (through the classification loss) and indirectly (through the combination loss).} 

The overall meta-learning approach is shown in Algorithm \ref{alg:meta}.

\begin{algorithm}[!ht]
\caption{{Forecast combination based on multi-task learning (DNN-MTL)}}
\label{alg:meta}
\tcc{OFFLINE PHASE: BUILD THE METADATA AND TRAIN THE MODEL}
\begin{itemize}[align=left, itemsep=0em, rightmargin=1em]
    \item[\textbf{Input}] Dataset of $N$ time series $\{\mathbf{s}^1, \dots, \mathbf{s}^N \}$, set of $M$ methods, forecasting horizon $H$.
    \item[\textbf{Output}] A function $f_{\textrm{meta}} \colon \mathbb{R}^T \to \mathbb{R}^M$ from a time series to a set of $M$ weights, one for each method.
\end{itemize}
\For{$i = 1:N$}{
\begin{enumerate}[itemsep=0em, leftmargin=1em, rightmargin=3em]
    \item Split $\mathbf{s}^i$ into training $\mathbf{x}^i = [y^i_{1}, \dots, y^i_{T}]^\top$ and test $\mathbf{y}^i = [y^i_{T+1}, \dots, y^i_{T+H}]^\top$ periods according to a temporal hold-out strategy (see Figure \ref{fig:holdout}).
    \item Fit each base forecasting method over the training period $\mathbf{x}^i$ and generate the matrix of forecasts ${\mathbf{F}}^i \in \mathbb{R}^{H \times M}$ over the test period $\mathbf{y}^i$. 
    \item Extract the similarity matrix $\mathbf{Q}^i \in \mathbb{R}^{M \times M}$ and the accuracy vector $\mathbf{c}^i \in \mathbb{R}^{M}$ over the test period $\mathbf{y}^i$ from the matrix of forecasting errors $\mathbf{E}^i$.
    \item Solve problem \eqref{prob:selection} and construct the ground-truth classification label $\mathbf{o}^i_{\textrm{cls}}$.
\end{enumerate}
}
Train $f_{\textrm{meta}}$ on the meta-data by solving
\begin{align*}
    \min_{\mathbf{\Theta}} \ \mathcal{L}(\mathbf{x}^i, \mathbf{y}^i, {\mathbf{F}}^i, \mathbf{o}^i_{\textrm{cls}}; \mathbf{\Theta}) 
\end{align*}

\tcc{ONLINE PHASE: FORECAST NEW TIME SERIES}
\begin{enumerate}[align=left, itemsep=0em]
    \item[\textbf{Input}] Trained meta-learner $f_{\textrm{meta}}$, dataset of $K$ time series $\{ \tilde{\mathbf{s}}^1, \dots,\tilde{\mathbf{s}}^K \}$.
    \item[\textbf{Output}] Combined forecasts.
\end{enumerate}
\For{$i = 1:K$}{
\begin{enumerate}[itemsep=0em, leftmargin=1em, rightmargin=3em]
    \item Compute weights $\tilde{\mathbf{w}}^i = f_{\textrm{meta}}(\tilde{\mathbf{s}}^i)$.
    \item Generate the forecasting matrix $\tilde{\mathbf{F}}^i \in  \mathbb{R}^{H \times M}$.
    \item Compute the combined forecasts $\tilde{\mathbf{y}}^i = \tilde{\mathbf{F}}^i \tilde{\mathbf{w}}^i$.
\end{enumerate}
}
\end{algorithm}

\subsection{Neural network architecture}
In this section, we describe the architectural components of the deep neural network. Our network is built on CNNs, which employ convolution operations to automatically extract features. They are comprised of a series of convolution layers and nonlinear activation functions, which are trained to identify valuable and complex features within the input data \citep{mancuso2021machine}. %As a result, CNNs are commonly used in various computer vision tasks, including object detection, image segmentation, and image classification \citep{li2021survey}. However, CNNs' advantages are not limited to visual data alone, as they are also used in applications that handle one-dimensional sequential data, such as time-series forecasting \citep{liu2019nonpooling, semenoglou2023image}. 

\begin{figure}[!ht]
    \centering
    \includegraphics[scale=0.38]{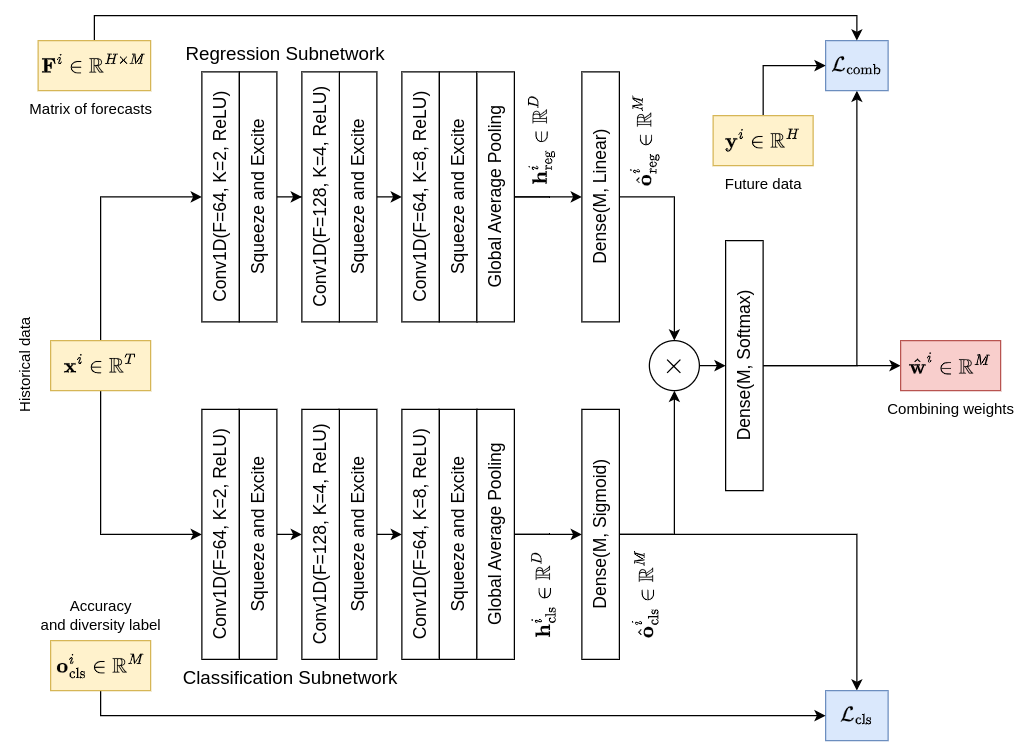}
    \caption{The network architecture designed to optimize the weights of the convex combination of base forecasting models for an input time series. In each convolutional layer, ``F'' and ``K'' denote the number of filters and the kernel size, respectively. {Input time series data and meta-data are depicted in yellow, the components of the loss function in blue, and the output of the meta-learner in red.}}
    \label{fig:network}
\end{figure}

Since the CNN architecture itself is not the main focus of our proposal, we employ for both subnetworks the same CNN architecture used by \cite{ma2021retail}.
To extract features, each subnetwork consists of three stacked temporal convolutional blocks. Each convolutional block comprises a convolutional layer and a ReLU activation function. Furthermore, convolutional blocks incorporate a squeeze and excite layer \citep{hu2018squeeze}, which uses global average pooling to generate summary statistics over the learned feature map, capturing contextual information outside each filter's focused feature of the input series. The excite operation in the squeeze and excite blocks introduces dynamic dependencies among the learned features, allowing the network to assign more importance to relevant features,  when needed. The ﬁlters of the three convolutional layers in each subnetwork are set to 64, 128, and 64 respectively, whereas the kernel sizes are set to 2, 4, and 8. As we can see in Figure \ref{fig:network}, our architecture consists of two identical and independent feature extractors with task-specific output branches.
In both subnetworks, the last temporal convolutional block is followed by a global average pooling layer to reduce network parameters. In the classification subnetwork, a dense layer with a sigmoid activation function is used to map the learned features into a set of labels, where the dimension matches the number of base forecasters. In contrast, the regression subnetwork outputs a set of unnormalized weights, also through a dense layer with linear activation function, with a dimension equal to the number of base forecasters.
To effectively leverage information between tasks, the unnormalized weights of the convex combination are then element-wise multiplied with the labels learned by the auxiliary task. The classification task allows the network to emphasize forecasting methods that are more accurate and diverse for the corresponding task, and downplay the effect of inaccurate and highly correlated ones. Both branches are then followed by a softmax layer that is trained to predict the final weighted combination of the forecasting methods. {As we will show in the numerical results, the auxiliary branch learns a representation of the input time series that differs from the representation learned by the regression branch. As a result, the multi-task architecture enables the model to capture distinct aspects of the data, combining information relevant for forecast accuracy with information related to diversity among forecasting methods.} Figure \ref{fig:network} illustrates the training phase by detailing the inputs and outputs of the meta-learner. The figure specifically shows how the various inputs, represented by historical data $\mathbf{x}^i$, ground-truth observations $\mathbf{y}^i$, and meta-data $\mathbf{F}^i$ and $\mathbf{o}^i_\textrm{cls}$ (highlighted in yellow), are fed into the meta-learner to produce the output weights $\mathbf{\hat{w}}^i$ (highlighted in red). It also depicts how the network processes these inputs through its blocks to generate intermediate results, as well as the components necessary for computing the addends $\mathcal{L}_\textrm{comb}$, $\mathcal{L}_\textrm{cls}$ of the custom loss function $\mathcal{L}$ (highlighted in blue). {From a computational perspective, unlike previous meta-learning approaches such as FFORMA and its variants, our method requires solving a small QP problem for each time series. These problems are convex and involve a limited number of variables, making their computational cost negligible relative to the overall runtime. In line with FFORMA and related approaches, the dominant cost arises in the offline phase, primarily due to fitting the base forecasting methods. The neural network can be trained on standard CPU hardware with moderate computational effort and can be further accelerated using GPUs. Finally, the online phase is computationally lightweight, as it only requires a forward pass through the neural network followed by a simple weighted combination of forecasts.}

\section{Experimental setup}\label{section:implementation}
\subsection{Time series datasets}
As the main benchmark to evaluate the forecasting accuracy of the proposed methodology, we use the M4 competition dataset. This dataset includes 100,000 time series of varying frequencies and is publicly available in the ``M4comp2018'' R package. We focus on the yearly, quarterly, and monthly series which represent 95\% of the competition’s series. {The limited number of available series for weekly, daily, and hourly frequencies suggested their exclusion from the analysis.}
The yearly subset includes 23,000 series with lengths ranging from 13 to 835 observations and with forecast horizons of 6 periods. The quarterly subset consists of 24,000
series with 8 forecast horizons and the series length ranges from 16 to 866 periods. Finally, the monthly subset contains 48,000 time series with a forecasting horizon of 18 periods ranging from 42 to 2,794 sample observations. Before feeding data to the neural network, all the input time series are standardized so that the mean of observed values is 0 and the standard deviation is 1. {Like other meta-leaning methods that use CNNs, the input series must be preprocessed to ensure they have the same length.} {Thus, for each frequency, time series are padded or truncated to a target length, determined as the median length rounded to the nearest power of 2.} Shorter series are padded with zeros, and longer series are truncated from the beginning. As a result, for yearly, quarterly, and monthly time series we consider lengths of 32, 64, and 128 observations, respectively. {Note that the base forecasters are fitted directly on the raw observations from the training period of each series without any truncating or padding operations.}

{To further evaluate the usefulness of the proposed forecasting framework in practice, we use the LargeST dataset \citep{liu2024largest}, a large-scale traffic flow time series dataset from the California Department of Transportation Performance Measurement System (PeMS) \citep{chen2001freeway}. PeMS provides real-time traffic data collected from 18,954 sensors across California at 5-minute intervals. {Following the experimental setup outlined by \cite{liu2024largest}, we select data from 8,600 sensors, each of which recorded data in both 2018 and 2019. Since each sensor corresponds to a time series, this results in a total of 17,200 time series. We aggregate this data and generate two datasets with different temporal granularities: daily and hourly frequencies. The length of the daily series has been truncated to 6 months (180 observations), while the hourly series have been truncated to 15 days (360 observations).}
All time series have the same length, so no pre-padding or truncation is required for input into the meta-learner. Similarly to the M4 dataset, all the input time series are standardized. The forecasting horizon is set to 7 days for the daily series and 24 hours for the hourly series.}

\subsection{Evaluation metrics}
We evaluate the forecasting accuracy using the Symmetric Mean Absolute Percentage Error (SMAPE) and the Mean Absolute Scaled Error (MASE). Consider $N$ time series $\left\{\mathbf{y}^1, \dots, \mathbf{y}^N\right\}$, where each individual series, denoted as $\mathbf{y}^i = \left[y^i_1, \dots, y^i_{T_i}\right]^\top$, has length $T_i$ and forecasting horizon $H_i$. The predicted values for the $i$-th series are collected in the vector $\hat{\mathbf{y}}^i = \left[\hat{y}^i_{T_i + 1}, \dots, \hat{y}^i_{T_i + H_i}\right]^\top$. Thus, for the $i$-th time series, the
$\textrm{SMAPE}^i$ is given by
\begin{equation}\label{smape}
    \textrm{SMAPE}^i = \textrm{SMAPE}(\mathbf{y}^i, {\hat{\mathbf{y}}}^i) = \frac{1}{H_i}\sum_{h=1}^{H_i}\frac{2\cdot\left|y^i_{T_i + h}-\hat{y}^i_{T_i+h}\right|}{\left|y^i_{T_i + h}\right|+\left|\hat{y}^i_{T_i+h}\right|}.
\end{equation}
The SMAPE is easy to interpret and has an upper bound of 2 when either
actual or predicted values are zero or when actual and predicted are opposite signs. On the other hand, the $\textrm{MASE}^i$ compares the forecast accuracy between a specific forecast algorithm and the naïve method which sets all forecasts to be the value of the last observation or period. It is defined as
\begin{equation}\label{mase}
    \textrm{MASE}^i = \textrm{MASE}(\mathbf{y}^i, \hat{\mathbf{y}}^i) = \frac{1}{H_i}\frac{\sum_{t=1}^{H_i} \left|y^i_{T_i + h}-\hat{y}^i_{T_i + h}\right|}{\frac{1}{T_i-s_i}\sum_{t=s_i+1}^{T_i}\left|y^i_{t}-y^i_{t-s_i}\right|},
\end{equation}
where $s_i$ is the frequency of the data. The numerator is the out-of-sample mean absolute error of the
method evaluated across the forecast horizon $H_i$, and the denominator is the
in-sample {absolute forecast error of the ``naïve forecasting method" with seasonal period $s_i$}. The $\textrm{SMAPE}^i$ and $\textrm{MASE}^i$ can be used to compare forecasting methods on a single
series and, because they are scale-free, to compare the forecasting accuracy across series. 

Moreover, we evaluate the forecasting accuracy by using the Overall Weighted Average (OWA) score which is the measure used in the M4 competition \citep{makridakis2020m4}. It is
based on SMAPE and MASE and it is given by
\begin{equation}\label{owa}
    \textrm{OWA} = \frac{1}{2} \frac{\sum_{i=1}^N \textrm{SMAPE}(\mathbf{y}^i, \hat{\mathbf{y}}^i)}{\sum_{i=1}^N \textrm{SMAPE}(\mathbf{y}^i, \hat{\mathbf{z}}^i)} + \frac{1}{2} \frac{\sum_{i=1}^N \textrm{MASE}(\mathbf{y}^i, \hat{\mathbf{y}}^i)}{\sum_{i=1}^N \textrm{MASE}(\mathbf{y}^i, \hat{\mathbf{z}}^i)},
\end{equation}
where $\hat{\mathbf{z}}^i$ is the vector obtained with the naïve method for the $i$-th time series from one to $H_i$ steps ahead. %OWA provides a score for the entire collection of time series that may mislead the evaluation. For this reason, 
Finally, we consider the series-level OWA (sOWA) for the $i$-th series, introduced by \cite{lichtendahl2020some} and defined as:
\begin{equation}\label{sowa}
        \textrm{sOWA}^i = \frac{1}{2} \frac{\textrm{SMAPE}(\mathbf{y}^i, \hat{\mathbf{y}}^i)}{\textrm{SMAPE}(\mathbf{y}^i, \hat{\mathbf{z}}^i)} + \frac{1}{2} \frac{\textrm{MASE}(\mathbf{y}^i, \hat{\mathbf{y}}^i)}{ \textrm{MASE}(\mathbf{y}^i, \hat{\mathbf{z}}^i)}.
\end{equation}

\subsection{{Meta-data generation}}
\label{sec:meta-data-generation}
\begin{figure}[!ht]
            \centering
            \includegraphics[scale=0.4]{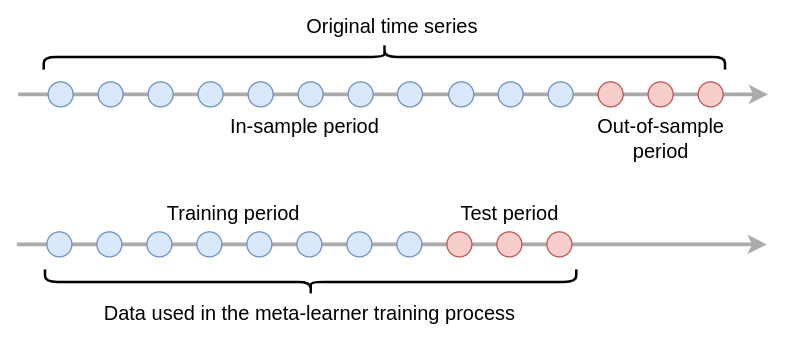}
            \caption{{The temporal hold-out strategy used to generate the training dataset.}}
            \label{fig:holdout}
        \end{figure}

{Following the related literature \citep{montero2020fforma, lichtendahl2020some, di2022sparse, kang2022forecast}, the training data for meta-learners is selected based on the temporal hold-out strategy applied to the in-sample period of all time series in each dataset (see Figure \ref{fig:holdout}).
More in detail, each original time series is divided into in-sample and out-of-sample periods. The length of the out-of-sample period is equal to the forecasting horizon $H$. The in-sample period of each series $\mathbf{s}^i$ is further divided into training $\mathbf{x}^i$ and test periods $\mathbf{y}^i$, with the length of the test equal to the forecasting horizon $H$. As described in offline phase of Algorithm \ref{alg:meta}, all the base forecasting methods are fitted using only the training period $\mathbf{x}^i$. Then, forecasts from each base forecaster are generated for the test period, gathered in the matrix of forecasts $\mathbf{F}^i$, and compared with the true observations of the test period $\mathbf{y}^i$, yielding the matrix of forecasting errors $\mathbf{E}^i$. Based on this matrix, classification labels $\mathbf{o}^i_{\textrm{cls}}$ are
obtained by solving Problem \eqref{prob:selection}. Here $\mathbf{Q}^i$ is the correlation matrix of the forecasting error ${\mathbf{E}}^i$, and
$c^i_j = -~\textrm{sOWA}^i_j$ for all $j \in \{1, \dots, M\}$, where $\textrm{sOWA}^i_j$ denotes the sOWA error reported by the $j$-th base forecasting method for the $i$-th series. 
We train the meta-learning model using as input only the observations in the training period $\mathbf{x}^i$, the ground-truth observations in the test period $\mathbf{y}^i$, the forecasts $\mathbf{F}^i$, and the labels $\mathbf{o}^i_{\textrm{cls}}$, which constitute the meta-data for each series. Observations from the out-of-sample period that have been held out at the beginning are exclusively used to evaluate the forecasts generated by the trained meta-learner.}

%{To determine the values of the hyperparameters, we employ a validation procedure based exclusively on the training portion of the data. More precisely, after applying the temporal hold-out strategy illustrated in Figure~\ref{fig:holdout}, the resulting data are further divided at the series level into training and validation subsets. Thus, the held-out out-of-sample periods used for the final evaluation are never used during hyperparameter selection, label generation, or model training.}

\subsection{Implementation details}
As candidates for base forecasters, we use the nine methods employed in previous studies ($M = 9$). These are described in Table \ref{tab:forecasting_methods} and implemented in the ``forecast" package in R \citep{hyndmanForecastPackage}.
\begin{table}[!ht]
    \centering
    \footnotesize
    \begin{tabular}{l|p{7cm}|l}
        \toprule
        \textbf{Method} & \textbf{Description} & \textbf{R function} \\
        \midrule
        \multirow{1}{*}{ARIMA} & Automated ARIMA algorithm \citep{hyndman2008automatic} & \texttt{auto.arima()} \\
        \multirow{1}{*}{ETS} & Automated exponential smoothing algorithm \citep{hyndman2002state} & \texttt{ets()} \\
        \multirow{1}{*}{NNETAR} & Feed-forward neural networks with a single hidden layer and AR inputs & \texttt{nnetar()} \\
        \multirow{1}{*}{TBATS} & TBATS model \citep{de2011forecasting} & \texttt{tbats()} \\
        \multirow{1}{*}{STLM} & Seasonal and trend decomposition using Loess with AR seasonally adjusted series \citep{cleveland1990stl} & \multirow{1}{*}{\texttt{stlm(modelfunction=`ar')}} \\
        \multirow{1}{*}{RW} & Random walk with drift & \texttt{rwf(drift=TRUE)} \\
        \multirow{1}{*}{THETA} & Theta method \citep{assimakopoulos2000theta} & \multirow{1}{*}{\texttt{thetaf()}} \\
        \multirow{1}{*}{NAIVE} & Naïve method (forecasting based on the most recent observation) & \texttt{naive()} \\
        \multirow{1}{*}{SNAIVE} & Seasonal naïve method (forecasting based on the most recent seasonal period) & \multirow{1}{*}{\texttt{snaive()}} \\
        \bottomrule
    \end{tabular}
    \caption{The nine individual forecasting methods of the pool, or {base learners}. The table provides the acronym of the method used throughout the paper (first column), the main reference where the method was introduced (second column), and the R function used for the experiments (third column).}
    \label{tab:forecasting_methods}
\end{table}
We train a distinct meta-learner for each group of series. Each meta-learner is trained by minimizing the custom loss function in Eq. \eqref{eq:loss} with Adam optimizer \citep{kingma2014adam} with an initial learning rate set to 0.001 and a batch size of 64 time series. Early stopping is used to avoid overfitting.

{To determine the values of the hyperparameters, we employ a validation procedure based exclusively on the training portion of the data. More precisely, after applying the temporal hold-out strategy illustrated in Figure~\ref{fig:holdout}, the resulting data are further divided at the series level into training and validation subsets. Thus, the held-out out-of-sample periods used for the final evaluation are never used during hyperparameter selection, label generation, or model training.}
The parameter $\lambda$ in the loss function is selected through grid search over the candidate values $\lambda \in \{1, 2, 5, 7, 10\}$, choosing the value that minimizes the validation loss. {Similarly, the thresholding parameter $\tau$ in \eqref{prob:selection} is selected via cross-validation over the candidate values $\tau \in \left\{\frac{0.5}{M},\; \frac{0.75}{M},\; \frac{1}{M},\; \frac{1.25}{M},\; \frac{1.5}{M},\; \frac{1.75}{M}\right\}$}. The selected values of $\lambda$ and $\tau$ for each meta-learner are reported in Table \ref{tab:opt_tau_lambda}.

%{Note that, for each value of $\tau$, the labels $\mathbf{o}_i^{cls}$ are recomputed exclusively using the corresponding training subset, and the meta-learner is subsequently trained on the training subset and evaluated on the validation subset. The final hyperparameter values are selected according to forecasting performance on the validation data}. 

\begin{table}[!ht]
\begin{minipage}[c]{0.5\textwidth}
\centering
    \begin{tabular}{l c c}
    \toprule
\multicolumn{3}{c}{\textbf{M4 Dataset}}\\
\midrule
\textbf{Frequency} & \textbf{$\lambda$} & \textbf{$\tau$} \\
\midrule
Yearly & 1 & 0.111 \\
Quarterly & 2 & 0.139 \\
Monthly & 5 & 0.167 \\
\bottomrule
\end{tabular}\hfill%
    \end{minipage}\qquad
\begin{minipage}[c]{0.5\textwidth}
\centering
\begin{tabular}{l c c}
\toprule
\multicolumn{3}{c}{\textbf{LargeST Dataset}}\\
\midrule
\textbf{Frequency} & \textbf{$\lambda$} & \textbf{$\tau$} \\
\midrule
Daily & 1 & 0.111 \\
Hourly & 7 & 0.167 \\
\bottomrule
\end{tabular}
\end{minipage}
\caption{{Selected values of $\lambda$ and $\tau$ for each dataset and series frequency.}}
\label{tab:opt_tau_lambda}
\end{table}

The neural network is written in Python 3.9 and implemented with TensorFlow 2.12. QP programs are solved using ``quadprog'' package in R. All the experiments are performed on a laptop with an Intel(R) i7-13700H CPU, 32 GB RAM, and Ubuntu 22.04 LTS. {On this hardware, the average time required to solve a single QP problem was approximately 2 milliseconds. Training the network for a single data frequency required between 2 and 6 minutes, depending on the number and length of the time series. For the largest M4 subset, namely the monthly series with 48,000 time series, the label-generation procedure, including the cross-validation over candidate values of $\tau$ and $\lambda$, required approximately 3 hours.} The source code is available through the following link: \texttt{https://github.com/antoniosudoso/dnn-mtl-comb}.

\section{Experimental results}\label{section:experimental}
We compare the point forecast performance of the proposed multi-task forecast combination approach, from now on referred to as DNN-MTL, against the following benchmark methods:
\begin{itemize}
    \item The simple average approach, where the forecasts from all nine methods in the forecasting pool are combined with equal weights (AVERAGE).
    \item {The constrained least squares regression approach \citep{conflitti2015optimal}, where the weights of all the nine methods in the forecasting pool are found by minimizing the in-sample mean squared combined error under the restriction that the weights sum up to one and are non-negative (CLS-REG).}
    \item The meta-learner introduced by \citep{montero2020fforma} that uses XGBoost to link hand-crafted statistical time series features with forecasting errors (FFORMA).
    \item The recent method proposed by \cite{kang2022forecast} employs XGBoost to connect diversity-based time series features with forecasting errors (FFORMA-DIV).
\end{itemize}
{Furthermore, we conduct an ablation study on DNN-MTL to assess the contribution of the auxiliary classification branch and to evaluate its effect within the multi-task architecture. In particular, we consider the following variants:
\begin{itemize}
    \item {DNN-REG}: a regression-only model obtained by removing the classification branch, thereby eliminating both the gating mechanism and the influence of QP-derived labels, and resulting in a single-task meta-learner based solely on the combination loss.
    \item {DNN-REG-DIV: a regression-only model that incorporates diversity into the regression objective through an explicit regularization term. This variant uses the same architecture as DNN-REG, but augments the forecast-combination loss with a diversity penalty based on the similarity matrix $\mathbf{Q}^i$. Specifically, the model is trained by minimizing
    \begin{align*}
\mathcal{L}_{\text{comb}} +
\gamma
\frac{1}{N}
\sum_{i=1}^{N}
(\hat{\mathbf{w}}^i)^\top \mathbf{Q}^i \hat{\mathbf{w}}^i,
    \end{align*}
    where $\gamma$ is a regularization hyperparameter. Minimizing the quadratic term penalizes assigning large weights to highly correlated forecasting methods, thereby encouraging diversity. We select $\gamma$ through grid search over the candidate values $\gamma \in \{0.01, 0.05, 0.1, 0.5, 1, 10\}$, choosing the value that minimizes the validation loss.}
    \item {DNN-MTL ($\lambda = 0$)}: a variant of the proposed architecture where the auxiliary classification loss is disabled. In this setting, the diversity-aware supervision is not enforced, while the gating mechanism is retained.
    \item {DNN-MTL}: the full multi-task model.
\end{itemize}
}
To evaluate the forecasting performance of each method we use the overall weighted average (OWA), the average series-level OWA (Avg sOWA), the average SMAPE (Avg SMAPE), and the average MASE (Avg MASE). Lower values for these metrics indicate better forecasting accuracy.

\subsection{{Experiments on M4 time series}}

%Moreover, we include the standard deviation (SD) to provide insight into the variability of accuracy across the series, offering a perspective on accuracy risk \citep{lichtendahl2020some}. 
We present numerical results for each group of series of the M4 dataset in Table \ref{tab:owa_frequency}. Our meta-learner DNN-MTL consistently outperforms other approaches in terms of {OWA, Avg sOWA, Avg SMAPE, and Avg MASE} surpassing the state-of-the-art meta-learner FFORMA-DIV, which also incorporates diversity. In contrast, the simple average combination consistently falls short when compared to all the meta-learners. {Notably, the constrained least squares regression approach performs slightly better than the simple average.} The ablation results provide insight into the contribution of the different components of the proposed architecture. Comparing DNN-REG and DNN-MTL ($\lambda = 0$), we observe that the latter achieves improved performance, indicating that the modulation of the regression outputs already provides a benefit by incorporating information from the auxiliary branch.
{As for DNN-REG-DIV, the results show that explicitly penalizing correlated forecasting methods generally improves the performance of the regression-only model, confirming the usefulness of diversity information in forecast combination. The best values of the regularization parameter $\gamma$ selected through validation were $0.1$, $0.5$, and $0.5$ for yearly, quarterly, and monthly series, respectively.}
However, the full multi-task formulation consistently yields the best results, highlighting the importance of the auxiliary classification loss in stabilizing and guiding the gating mechanism. Overall, the full model outperforms its ablated variants, showing that the observed gains arise from the combined effect of gating and diversity-aware supervision.

\begin{figure}[!ht]%
    \centering
    \includegraphics[scale=0.52]{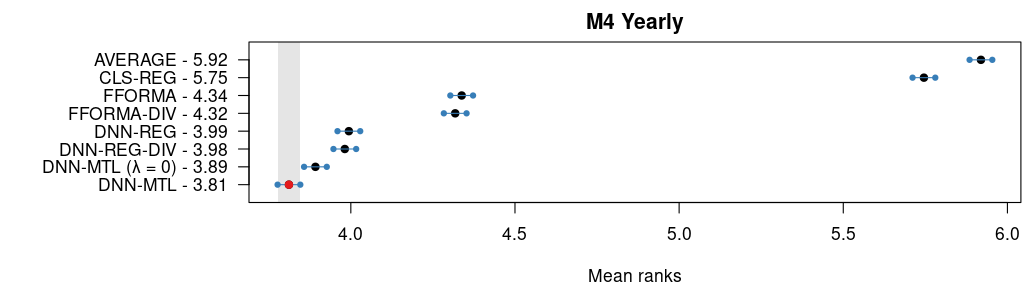}
    \includegraphics[scale=0.52]{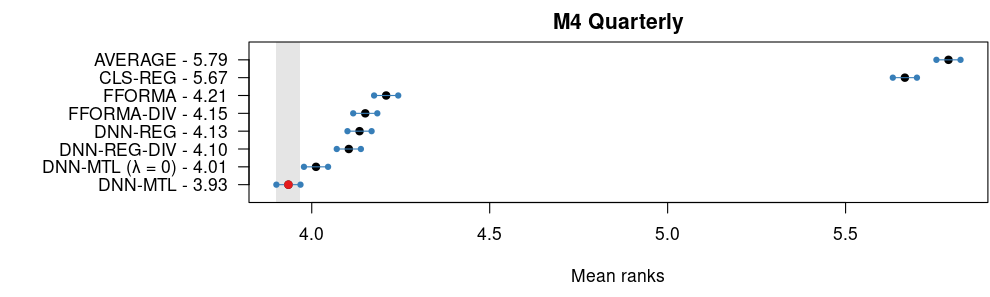}
    \includegraphics[scale=0.52]{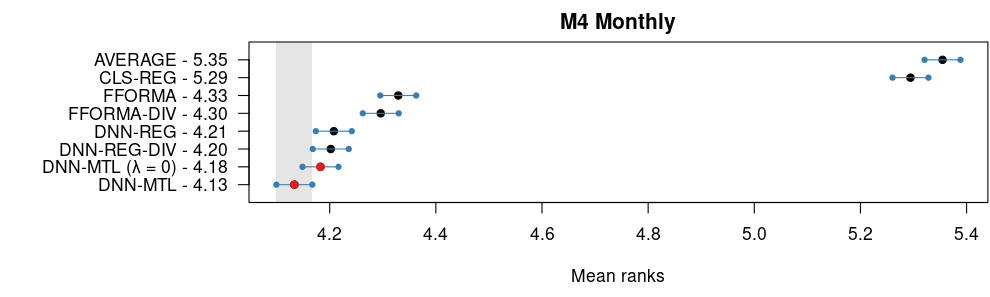}
    \caption{MCB Nemenyi test results, average ranks, and 95\% confidence intervals for yearly, quarterly, and monthly time series of the M4 dataset. Forecast combination methods are sorted vertically according to the sOWA mean rank. The mean rank of each approach is shown to the right of its name.} %Statistical differences in performance are observed if the intervals of two forecast combination procedures do not overlap.}%
    \label{fig:stat_test}%
\end{figure}

To formally test whether the performances among the considered methods are statistically different, we employ the non-parametric Friedman test and post-hoc Multiple Comparisons with the Best (MCB) Nemenyi test \citep{koning2005m3}, as implemented in the R package ``tsutils'' \citep{tsutils}. As outlined by \cite{kourentzes2019cross}, the Friedman test initially assesses whether at least one of the methods significantly differs from the others. If such a difference exists, the Nemenyi test is applied to identify groups of forecasting methods that do not exhibit statistically significant differences. This testing approach offers the advantage of not imposing any assumptions about the distribution of data and avoids the need for multiple pairwise comparisons between forecasts, which could bias the test results. We apply these tests on each data frequency based on the sOWA errors as shown in Figure \ref{fig:stat_test}. One can interpret the results in the following manner: lower average ranks indicate better performance, but there are no significant performance differences between any two methods if their confidence intervals overlap. According to Figure \ref{fig:stat_test} we observe that: (i) although FFORMA-DIV outperforms FFORMA on average, their differences are not statistically significant. These findings align with those documented in the recent study by \cite{kang2022forecast}; (ii) our method, DNN-MTL, takes the top position and generates forecasts that exhibit statistically significant differences when compared to those produced by other meta-learners.

\begin{table}[!ht]
%\footnotesize
    \centering
    \begin{tabular}{l c c c c}
    \toprule
& \multicolumn{3}{c}{\textbf{M4 Yearly}} &\\
\midrule
\multirow{2}{*}{\textbf{Method}} & \multirow{2}{*}{\textbf{OWA}} & \multicolumn{1}{p{1.50cm}}{\centering \textbf{Avg} \\ \textbf{sOWA}} &  \multicolumn{1}{p{1.50cm}}{\centering \textbf{Avg} \\ \textbf{SMAPE}} & \multicolumn{1}{p{1.50cm}}{\centering \textbf{Avg} \\ \textbf{MASE}} \\
\midrule 
      AVERAGE     & 0.949 & 1.204 & 0.153 & 3.647 \\ 
      CLS-REG     & 0.946 & 1.193 & 0.151 & 3.532 \\
      FFORMA      & 0.799 & 0.996 & 0.135 & 3.085 \\
      FFORMA-DIV  & 0.798 & 0.979 & 0.129 & 3.071 \\
      {DNN-REG}     &  {0.801} &  {0.896} &  {0.128} &  {3.080} \\
      {DNN-REG-DIV} & {0.799} & {0.896} & {0.128} & {3.071} \\
      {DNN-MTL ($\lambda = 0$)} & {0.800} & {0.879} & {0.118} & {3.057} \\
      {DNN-MTL} & {\bftab{0.794}} & {\bftab{0.874}} & {\bftab{0.112}} & {\bftab{3.025}}\\
\end{tabular}%

    \begin{tabular}{l c c c c}
    \toprule
& \multicolumn{3}{c}{\textbf{M4 Quarterly}} &\\
\midrule
\multirow{2}{*}{\textbf{Method}} & \multirow{2}{*}{\textbf{OWA}} & \multicolumn{1}{p{1.50cm}}{\centering \textbf{Avg} \\ \textbf{sOWA}} &  \multicolumn{1}{p{1.50cm}}{\centering \textbf{Avg} \\ \textbf{SMAPE}} & \multicolumn{1}{p{1.50cm}}{\centering \textbf{Avg} \\ \textbf{MASE}} \\
\midrule
      AVERAGE     & 0.916  & 0.955 & 0.125 & 1.243 \\ 
      CLS-REG     & 0.912  & 0.949 & 0.120 & 1.239 \\
      FFORMA      & 0.847  & 0.910 & 0.097 & 1.112 \\ 
      FFORMA-DIV  & 0.842  & 0.899 & 0.095 & 1.109 \\
      {DNN-REG}     & {0.845}  & {0.902} & {0.095} & {1.110} \\
    {DNN-REG-DIV} & {0.842} & {0.895} & {0.095} & {1.108} \\
      {DNN-MTL ($\lambda = 0$)} & {0.840} & {0.890} & {0.091} & {1.092}\\
      {DNN-MTL} & {\bftab{0.833}} & {\bftab{0.886}} & {\bftab{0.085}} & {\bftab{1.090}} \\
\end{tabular}%

\begin{tabular}{l c c c c}
    \toprule
& \multicolumn{3}{c}{\textbf{M4 Monthly}} &\\
\midrule
\multirow{2}{*}{\textbf{Method}} & \multirow{2}{*}{\textbf{OWA}} & \multicolumn{1}{p{1.50cm}}{\centering \textbf{Avg} \\ \textbf{sOWA}} &  \multicolumn{1}{p{1.50cm}}{\centering \textbf{Avg} \\ \textbf{SMAPE}} & \multicolumn{1}{p{1.50cm}}{\centering \textbf{Avg} \\ \textbf{MASE}} \\
\midrule
      AVERAGE     & 0.911 & 0.952 & 0.136 & 0.983 \\ 
      CLS-REG     & 0.909 & 0.948 & 0.135 & 0.983 \\
      FFORMA      & 0.858 & 0.905 & 0.129 & 0.898 \\
      FFORMA-DIV  & 0.851 & 0.904 & 0.128 & 0.891 \\
      {DNN-REG}     & {0.862} & {0.909} & {0.128} & {0.892} \\
    {DNN-REG-DIV} & {0.855} & {0.901} & {0.126} & {0.889} \\
    {DNN-MTL ($\lambda = 0$)} & {0.853} & {0.888} & {0.123} & {0.885} \\
      {DNN-MTL} & {\bftab{0.846}} & {\bftab{0.887}} & {\bftab{0.112}} & {\bftab{0.879}} \\
\bottomrule
    \end{tabular}
\caption{{Test set evaluation metrics for yearly, quarterly, and monthly time series of the M4 dataset. OWA, average (Avg) sOWA, SMAPE, and MASE for each combination method.} Lower values correspond to better forecasting accuracy. The best-performing method is highlighted in bold.}
    \label{tab:owa_frequency}
\end{table}

\subsection{{Experiments on LargeST time series}}
\begin{figure}[!ht]
    \centering
    \includegraphics[scale=0.55]{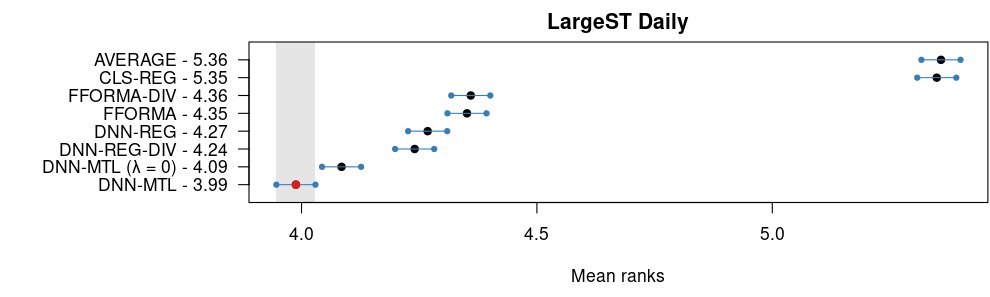}
    %\vspace{0.1cm}
    \includegraphics[scale=0.55]{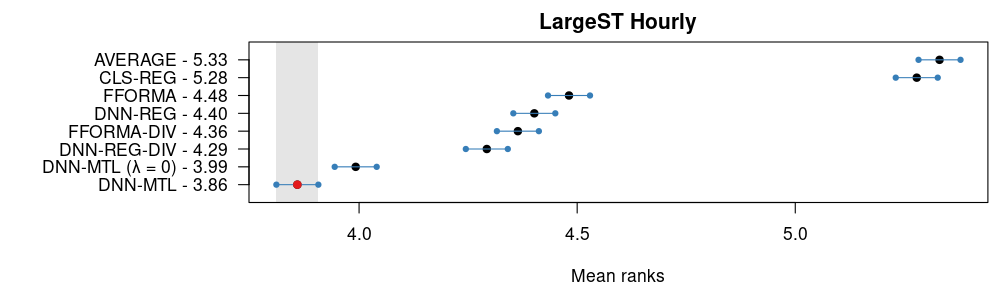}
    \caption{{MCB Nemenyi test results, average ranks, and 95\% confidence intervals for daily and hourly time series of the LargeST dataset. Forecast combination methods are sorted vertically according to the sOWA mean rank. The mean rank of each approach is shown to the right of its name.}}%Statistical differences in performance are observed if the intervals of two forecast combination procedures do not overlap.}
    \label{fig:stat_test_traffic}
\end{figure}

{Computational results for each group of series of the LargeST dataset are presented in Table \ref{tab:owa_frequency_largeST}. 
Similarly to the results observed in the M4 series, our method DNN-MTL outperforms all the other meta-learners in terms of OWA, Avg sOWA, Avg SMAPE, and Avg MASE. {Moreover, the ablation experiment confirms that DNN-MTL produces better forecasts than its ablated variants}. {The baseline DNN-REG-DIV leads to similar conclusions on the LargeST dataset. The best values of the regularization parameter $\gamma$ selected through validation were $0.05$ and $0.1$ for daily and hourly series, respectively.}

Figure \ref{fig:stat_test_traffic} illustrates the MCB Nemenyi test results conducted on each data frequency based on the sOWA errors. In this figure, our DNN-MTL approach consistently secures the top position, generating statistically different forecasts compared to other meta-learners. Although the CLS-REG method does not outperform the meta-learning methods, it exhibits enhanced performance relative to the simple average. Despite this improvement, statistically significant differences in forecast accuracy are not evident for daily series.
In conclusion, numerical results on the LargeST dataset demonstrate the robustness and applicability of our proposed forecasting framework in real-world applications.}

\begin{table}[!ht]
%\footnotesize
    \centering
    \begin{tabular}{l c c c c}
    \toprule
& \multicolumn{3}{c}{\textbf{LargeST Daily}} &\\
\midrule
\multirow{2}{*}{\textbf{Method}} & \multirow{2}{*}{\textbf{OWA}} & \multicolumn{1}{p{1.50cm}}{\centering \textbf{Avg} \\ \textbf{sOWA}} &  \multicolumn{1}{p{1.50cm}}{\centering \textbf{Avg} \\ \textbf{SMAPE}} & \multicolumn{1}{p{1.50cm}}{\centering \textbf{Avg} \\ \textbf{MASE}} \\
\midrule 
      AVERAGE & 1.267 & 1.459 & 0.121 & 3.457 \\ 
      CLS-REG & 1.017 & 1.205 & 0.115 & 2.279 \\ 
      FFORMA & 0.804  & 0.843 & 0.090 & 1.017 \\ 
      FFORMA-DIV & 0.803 & 0.842 & 0.090 & 1.015 \\ 
          {DNN-REG} & {0.792} & {0.839} & {0.085} & {0.866} \\ 
      {DNN-REG-DIV} & {0.790} & {0.839} & {0.085} & {0.865} \\
      {DNN-MTL ($\lambda = 0$)} & {0.773} & {0.829} & {0.083} & {0.861} \\
      {DNN-MTL} & {\bftab{0.771}} & {\bftab{0.820}} & {\bftab{0.079}} & {\bftab{0.859}} \\ 
\end{tabular}%

    \begin{tabular}{l c c c c}
    \toprule
& \multicolumn{3}{c}{\textbf{LargeST Hourly}} &\\
\midrule
\multirow{2}{*}{\textbf{Method}} & \multirow{2}{*}{\textbf{OWA}} & \multicolumn{1}{p{1.50cm}}{\centering \textbf{Avg} \\ \textbf{sOWA}} &  \multicolumn{1}{p{1.50cm}}{\centering \textbf{Avg} \\ \textbf{SMAPE}} & \multicolumn{1}{p{1.50cm}}{\centering \textbf{Avg} \\ \textbf{MASE}} \\
\midrule
      AVERAGE & 0.975 & 1.105 & 0.224 & 1.296\\ 
      CLS-REG & 0.507 & 0.561 & 0.221 & 0.852\\ 
      FFORMA & 0.483 & 0.496 & 0.184 & 0.656\\ 
      FFORMA-DIV & 0.478 & 0.481 & 0.166 & 0.635\\ 
          {DNN-REG} & {0.422} & {0.433} & {0.114} & {0.536}\\ 
      {DNN-REG-DIV} & {0.421} & {0.433} & {0.111} & {0.532} \\
      {DNN-MTL ($\lambda = 0$)} & {0.414} & {0.433} & {0.103} & {0.504} \\
      {DNN-MTL} & {\bftab{0.411}} & {\bftab{0.425}} & {\bftab{0.096}} & {\bftab{0.502}}\\ 
\bottomrule
\end{tabular}
\caption{{Test set evaluation metrics for daily and hourly time series of the LargeST dataset. OWA, average (Avg) sOWA, SMAPE, and MASE for each combination method. Lower values correspond to better forecasting accuracy. The best-performing method is highlighted in bold.}}
    \label{tab:owa_frequency_largeST}
\end{table}

\clearpage

\subsection{Analysis of learned representations}
\label{sec:representation_analysis}
{To analyze the features learned by the regression and classification branches, we study the similarity between the latent representations extracted by the two subnetworks. For each time series, we collect the 64-dimensional feature vectors produced by the regression and classification branches, resulting in two matrices $\mathbf{H}_{\mathrm{reg}}, \mathbf{H}_{\mathrm{cls}} \in \mathbb{R}^{N \times 64}$.}

{We first quantify the similarity between the two representation spaces using linear Centered Kernel Alignment (CKA), a widely used metric for comparing neural network representations \citep{kornblith2019similarity}. CKA is invariant to orthogonal transformations and isotropic scaling, making it suitable for comparing features learned by the two subnetworks. {The CKA score takes values in $[0,1]$, where values close to 1 indicate highly similar representations, while values close to 0 indicate limited representational similarity between the learned feature spaces.}

{To further investigate the relationship between the learned features, we perform a cross-correlation analysis. Specifically, after standardizing each feature dimension, we compute the pairwise Pearson correlation between all features across the two representations, resulting in a $64 \times 64$ correlation matrix. From this matrix, we report the mean absolute correlation, denoted as Mean $|\rho|$, which provides a summary measure of feature-level alignment between the two branches.}

{Table \ref{tab:representation_similarity} reports the CKA scores and mean absolute correlations for each dataset and frequency. The results consistently show that the CKA values lie in the range $[0.28, 0.35]$, indicating low similarity between $\mathbf{H}_{\mathrm{reg}}$ and $\mathbf{H}_{\mathrm{cls}}$. This suggests that the two branches do not learn equivalent encodings, but instead capture partially overlapping yet distinct information from the input time series.}
{The cross-correlation analysis provides further insight into this behavior}. The correlation structure does not exhibit a clear one-to-one alignment between features of the two branches. Instead, correlations are distributed across multiple feature pairs, indicating that the representations are not simple permutations or linear transformations of each other. This observation is also reflected in the relatively low values of Mean $|\rho|$, which remain well below one across all settings.

Thus, the CKA and cross-correlation analyses jointly suggest that the regression and classification subnetworks do not learn redundant representations. While some shared structure is present, the two branches emphasize different aspects of the input data, which is consistent with their respective objectives.

\begin{table}[!ht]
\begin{minipage}[c]{0.5\textwidth}
\centering
\begin{tabular}{l c c}
\toprule
\multicolumn{3}{c}{\textbf{M4 Dataset}}\\
\midrule
\textbf{Frequency} & \textbf{CKA} & \textbf{Mean $|\rho|$} \\
\midrule
Yearly    & 0.28 & 0.21 \\
Quarterly & 0.32 & 0.24 \\
Monthly   & 0.35 & 0.26 \\
\bottomrule
\end{tabular}\hfill%
\end{minipage}\qquad
\begin{minipage}[c]{0.5\textwidth}
\centering
\begin{tabular}{l c c}
\toprule
\multicolumn{3}{c}{\textbf{LargeST Dataset}}\\
\midrule
\textbf{Frequency} & \textbf{CKA} & \textbf{Mean $|\rho|$} \\
\midrule
Daily  & 0.31 & 0.22 \\
Hourly & 0.34 & 0.25 \\
\bottomrule
\end{tabular}
\end{minipage}
\caption{{Representation similarity between the regression and classification branches across datasets and frequencies. CKA denotes the linear centered kernel alignment, while Mean $|\rho|$ represents the average absolute cross-feature correlation.}}
\label{tab:representation_similarity}
\end{table}

\subsection{Visual explanations}
%The interpretation of the features derived from CNNs frequently presents difficulties.
{We perform additional analysis to verify that the different base forecasting methods leverage distinct portions of the information in the time series. This qualitative information appears to indicate that the network behaves according to its design}. The analysis is based on Gradient-weighted Class Activation Mapping (Grad-CAM) \citep{selvaraju2017grad}, a method that enhances the interpretability of deep neural networks with visual heatmaps and highlights the regions of the input that mostly influence the network's decision for a particular class. 
We apply the method to analyze the behavior of the classification subnetwork. First, we disconnect,  then use it to compute,  for an input time series, the probability that a forecasting method is selected. 
{The method thus identifies those timesteps of the series that contributed the most to the selection of a specific forecasting method}. Overlaying the Grad-CAM heatmaps onto the original time series, we can visually verify the timesteps of the series that are motivating the selection of a method. 
{It is important to note that this information does not report on the relative importance of that method in producing the final forecast - this would derive also from the weights used to combine the methods - but only shows what are the timesteps under focus to make the model selection decisions.}
For space reasons, we report on Grad-CAM results for a small sample of the available data, considering yearly, quarterly, and monthly time series, with a threshold of 0.5 to convert each predicted probability into a 0-1 value from which the gradient of the predicted class score is computed. 
In Figures \ref{fig:gradcam_yearly}--\ref{fig:gradcam_monthly}, for each base forecasting method selected by the network, we analyze the heatmap produced for some examples of yearly, quarterly, and monthly time series, respectively. At each timestep, an importance score ranging from 0 to 1 is assigned. A value close to 1 indicates high significance of the corresponding timestep, while values near 0 indicate timesteps with low importance for the classification outcome. 
Looking at the figures, the heatmaps exhibit diversity in terms of the temporal regions they focus on within the input time series. These findings imply that the neural network is not merely relying on a single common pattern present in the time series but is considering multiple relevant segments. 
%
%{Indeed, the heatmap refers to the relevance of the timestep for the selection of the forecasting method, and only indirectly may point to the relative importance of the method in the final prediction.}
%
Furthermore, these distinctive areas of interest indicate that the network is leveraging different {timesteps} for different methods, likely reflecting the unique characteristics and strengths of each base forecaster. Finally, it is important to note that while the network is trained to learn both accurate and diverse methods, certain regions with high heatmap values appear to be shared among multiple methods. {These common regions indicate the particular temporal segments that hold significance in the context of more than one forecasting technique.}

\begin{figure}[!ht]%
    \centering
    \subfloat{{\includegraphics[scale=0.51]{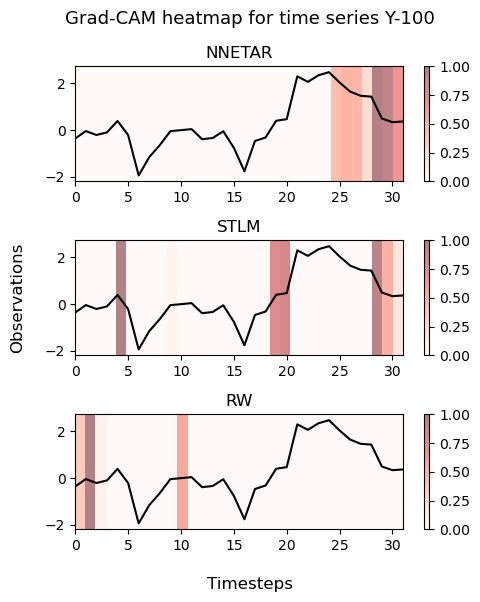} }}
    \subfloat{{\includegraphics[scale=0.51]{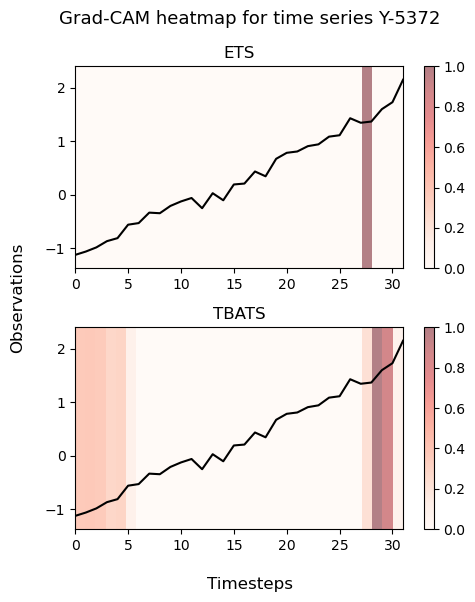} }}
    
    \subfloat{{\includegraphics[scale=0.51]{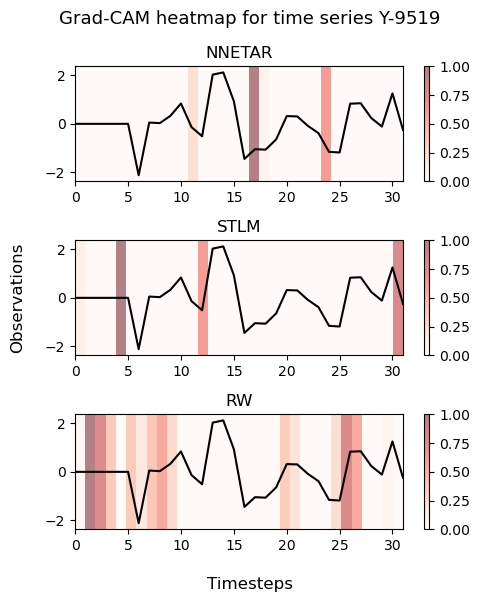} }}
    \subfloat{{\includegraphics[scale=0.51]{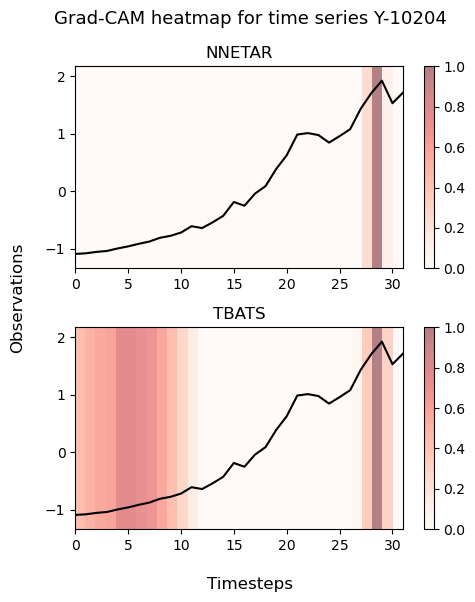} }}
        \caption{Grad-CAM visual explanations of the predicted base learners for a sample of 4 yearly test series. Time series have been normalized to zero mean and unit variance. For an input series, there is a heatmap for each forecasting method selected by the classification subnetwork.}
    \label{fig:gradcam_yearly}
\end{figure}

\begin{figure}[!ht]%
    \centering
    \subfloat{{\includegraphics[scale=0.51]{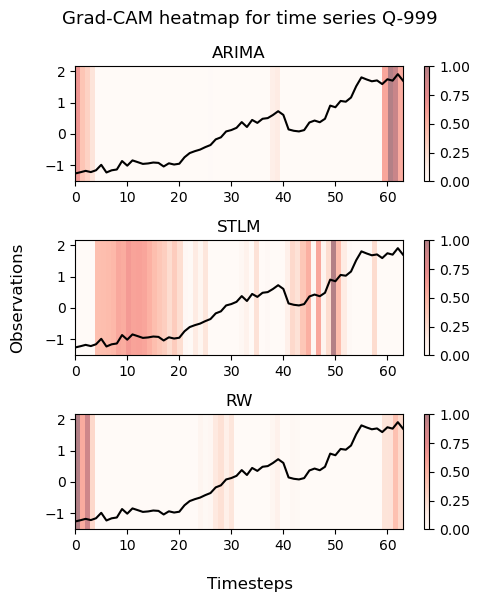} }}
    \subfloat{{\includegraphics[scale=0.51]{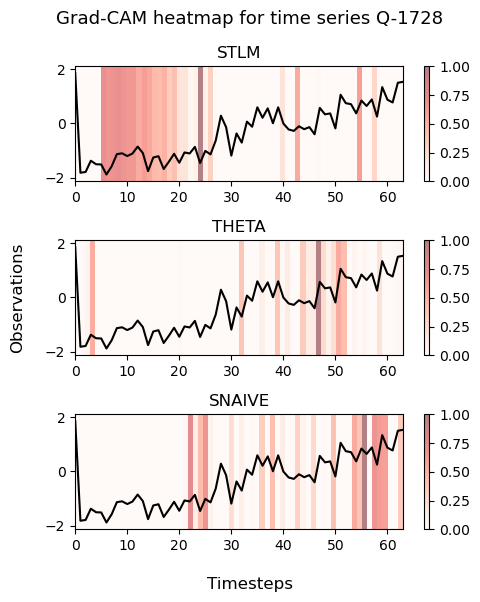} }}
    
    \subfloat{{\includegraphics[scale=0.51]{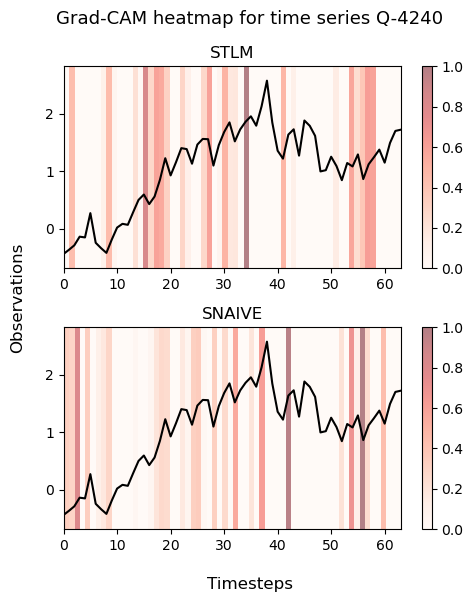} }}
    \subfloat{{\includegraphics[scale=0.51]{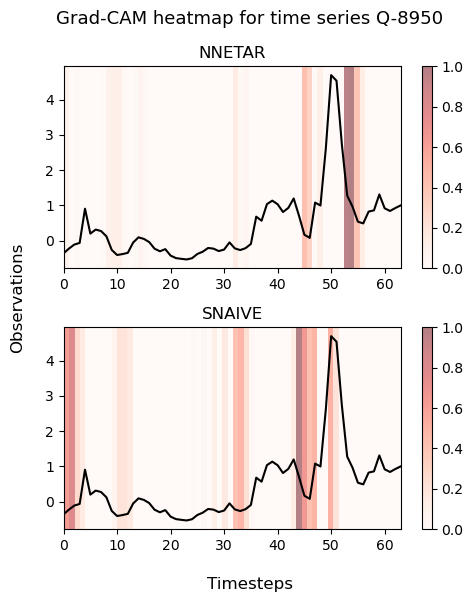} }}
        \caption{Grad-CAM visual explanations of the predicted base learners for a sample of 4 quarterly test series. Time series have been normalized to zero mean and unit variance. For an input series, there is a heatmap for each forecasting method selected by the classification subnetwork.}
    \label{fig:gradcam_quarterly}
\end{figure}

\begin{figure}[!ht]%
    \centering
    \subfloat{{\includegraphics[scale=0.51]{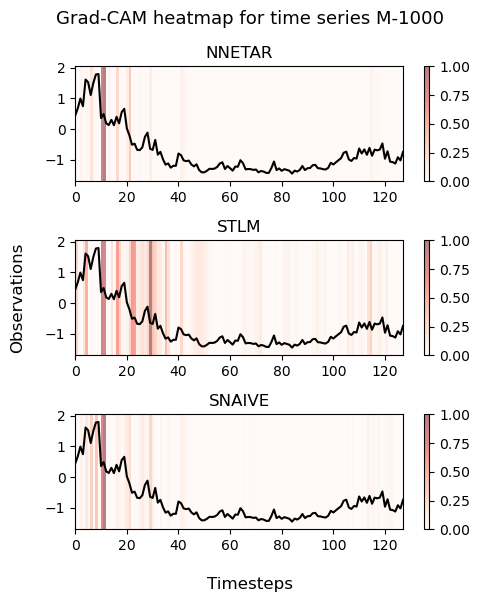} }}
    \subfloat{{\includegraphics[scale=0.51]{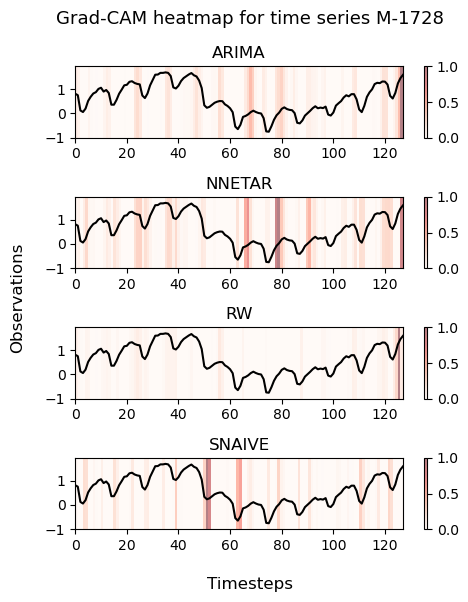} }}
    
    \subfloat{{\includegraphics[scale=0.51]{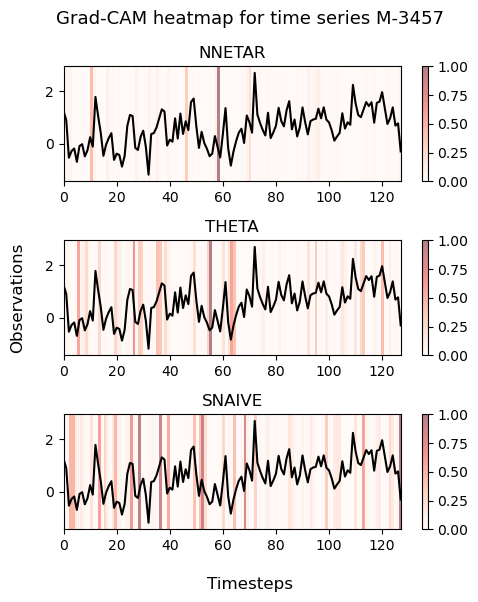} }}
  \subfloat{{\includegraphics[scale=0.51]{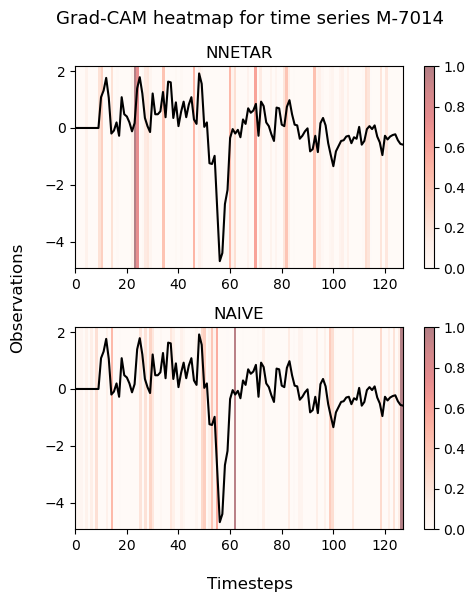} }}
    \label{fig:aaa}%
        \caption{Grad-CAM visual explanations of the predicted base learners for a sample of 4 monthly test series. Time series have been normalized to zero mean and unit variance. For an input series, there is a heatmap for each forecasting method selected by the classification subnetwork.}
    \label{fig:gradcam_monthly}
\end{figure}

\section{Conclusion}\label{section:conclusion}
We present a multi-task optimization methodology to improve the performances of convex combinations of forecasting models. Building on the literature, we use meta-learning to link the features of time series with the forecasts provided by a pool of base learners. Our meta-learner simultaneously addresses two related tasks within a multi-task paradigm.
{The main task is forecast combination, while an auxiliary task provides diversity-aware supervision during training.}
To accomplish this, we employ a deep neural network architecture with two branches to jointly solve the associated regression and classification problems. {The outputs of the two branches are then combined to obtain the final weights of the convex combination, with the auxiliary branch acting as a gating mechanism on the regression outputs}.
To provide labels for the classification task, we introduce an optimization-driven approach to identify the most appropriate method for a given time series, considering accuracy and diversity among the methods in the pool. 
Experimental results on the M4 competition dataset {and on a large-scale traffic flow dataset} demonstrate that this approach enhances the accuracy of point forecasts compared to state-of-the-art meta-learning methods.
%
%Moreover, gradient-based visual explanations provide interesting insights into discriminative regions in the input series that contribute to the network’s selection of a specific forecasting method.
%
Our approach presents an automated and adaptable tool for optimizing forecasting procedures, featuring the following key advantages. First, it relieves forecasters from the burden of filtering the methods according to diversity and accuracy criteria when conducting feature-based forecasting, as it effectively learns diversity information during training. Second, it can accommodate forecasts from a variety of methods, including statistical techniques, nonlinear approaches, and judgment-based forecasting. 
{Third, it can be readily applied to other forecasting domains, such as finance, weather, or healthcare. In particular, the proposed framework is not tied to a specific application setting, as it only requires a pool of candidate forecasting models and historical time series data.}
The proposed method is limited to producing only point forecasts; this motivates future research in the direction of adapting it to output interval forecasts. Another potential direction for future research is the extension to probability density forecasting \citep{hall2007combining}. In this context, meta-learning could play a role in generating weighted combinations of forecast distributions from various models.
We finally note that the proposed method identifies the same combining weights (and selected methods) for each step ahead of the forecasting horizon. To allow the weights to differ for different forecasting horizons, an alternative approach could involve training separate meta-learners for each horizon. Future research could explore the resulting trade-off between computational time and forecasting accuracy.

\section*{Declarations}
\subsection*{Funding}
The work presented in this paper has been supported by PNRR MUR project PE0000013-{FAIR} and CNR DIT.AD106.097 project UISH - Urban Intelligence Science Hub.

\subsection*{Competing Interests}
The authors have no financial or non-financial interests to disclose.
\subsection*{Ethical approval}
This article does not contain any studies with human participants or animals performed by any of the authors.

\begin{appendices}

\section{Details on the gradient computation}\label{appendix:gradient}
We perform some additional analysis to better understand the impact of $\mathcal{L}_{\textrm{comb}}$ on the gradient. To compute the gradient of the combined loss with respect to the $k$-th parameter $\Theta_k \in \mathbf{\Theta}$ of the network, we can follow the chain rule of differentiation. We rewrite the softmax function as
\begin{align*}
    \hat{w}^i_j = \frac{\textrm{exp}(\hat{z}^i_j)}{\sum_{t=1}^M \textrm{exp}(\hat{z}^i_t)}, \quad \forall j \in \{1, \dots, M\},
\end{align*}
where $\hat{z}^i_j = ( \hat{\mathbf{o}}^i_{\textrm{reg}} \odot  \hat{\mathbf{o}}^i_{\textrm{cls}})_j$.
Then, we have
\begin{align}\label{eq:full_gradient}
\frac{\partial \mathcal{L}_{\textrm{comb}}}{\partial \Theta_k} = \sum_{i=1}^N
    \frac{\partial \mathcal{L}^i_{\textrm{comb}}}{\partial \Theta_k} = \sum_{i=1}^N \sum_{j=1}^M \frac{\partial \mathcal{L}^i_{\textrm{comb}}}{\partial  \hat{{w}}^i_j} \sum_{t=1}^M \frac{\partial \hat{w}^i_j}{\partial \hat{z}^i_t} \cdot \frac{\partial \hat{z}^i_t}{\partial \Theta_k}.
\end{align}
%For the sake of simplicity, we focus on the contribution of the $i$-th time series to the loss function. Then
%Then, we have
%\begin{align*}
%    \frac{\partial \hat{w}^i_j}{\partial \Theta_k} &= %\frac{\partial \hat{w}^i_j}{\partial \hat{z}^i_t} %\cdot \frac{\partial \hat{z}^i_t}{\partial \Theta_k}.
%\end{align*}
One can verify that the derivative terms are given by:
\begin{align}
\frac{\partial \mathcal{L}^i_{\textrm{comb}}}{\partial  \hat{{w}}^i_j} &= \frac{1}{N \big\| \frac{1}{M} {\mathbf{F}}^i \mathbf{1}_M - \mathbf{y}^i \big\|_1} \sum_{h=1}^H  \textrm{sgn}\left(\sum_{k=1}^M {F}^i_{hk} \hat{w}^i_k - y^i_h \right) {F}^i_{hj},  \label{eq:1}\\
\frac{\partial \hat{w}^i_j}{\partial \hat{z}^i_t} &= \hat{w}^i_j (\delta_{jt} - \hat{w}^i_t),  \label{eq:2}\\
\frac{\partial \hat{z}^i_t}{\partial \Theta_k} &= \frac{\partial (\hat{\mathbf{o}}^i_{\textrm{reg}})_t}{\partial \Theta_k} \cdot (\hat{\mathbf{o}}^i_{\textrm{cls}})_t + (\hat{\mathbf{o}}^i_{\textrm{reg}})_t \cdot \frac{\partial (\hat{\mathbf{o}}^i_{\textrm{cls}})_t}{\partial \Theta_k}  \label{eq:3},
\end{align}
where $\delta_{jt}$ is the Kronecker delta and $\textrm{sgn}(\cdot)$ is the sign function.
Therefore, after plugging the above equations in Eq. \eqref{eq:full_gradient}, it can be seen that the individual outputs of regression and classification problems contribute to the parameter updates of the overall network through backpropagation.

\end{appendices}


\bibliography{sn-bibliography}% common bib file

\end{document}